\documentclass[manuscript]{acmart}

\AtBeginDocument{%
  \providecommand\BibTeX{{%
    \normalfont B\kern-0.5em{\scshape i\kern-0.25em b}\kern-0.8em\TeX}}}

\setcopyright{acmlicensed}
\acmJournal{TECS}
\acmYear{2024} \acmVolume{1} \acmNumber{1} \acmArticle{1} \acmMonth{1}\acmDOI{10.1145/3665278}

\usepackage{booktabs}
\usepackage{graphicx}
\usepackage{orcidlink}
\usepackage{caption}
\usepackage{subcaption}
\usepackage{fancyvrb}
\usepackage{breakurl}
\usepackage{xurl}
\usepackage{amsmath,amsfonts,upquote,textcomp,multirow}

\newcommand{\upquote}{\text{\textquotesingle}}
\usepackage{algorithm}
\usepackage{algcompatible}

\algdef{SE}[SUBALG]{Indent}{EndIndent}{}{\algorithmicend\ }
\algtext*{Indent}
\algtext*{EndIndent}

\makeatletter
\newcommand*{\rom}[1]{\expandafter\@slowromancap\romannumeral #1@}
\makeatother

\begin{document}

\title{On-device Online Learning and Semantic Management of TinyML Systems}

\author{Haoyu Ren}
\email{haoyu.ren@siemens.com}
\orcid{0000-0002-0241-6507}
\affiliation{%
  \institution{Siemens AG,}
  \institution{Technical University of Munich}
  \city{Munich}
  \state{Bavaria}
  \country{Germany}
}

\author{Darko Anicic}
\email{darko.anicic@siemens.com}
\orcid{0000-0002-0583-4376}
\affiliation{%
  \institution{Siemens AG}
  \city{Munich}
  \state{Bavaria}
  \country{Germany}
}

\author{Xue Li}
\email{xueli@eecs.uq.edu.au}
\orcid{0000-0002-4515-6792}
\affiliation{%
  \institution{The University of Queensland}
  \city{Brisbane}
  \state{Queensland}
  \country{Australia}
}

\author{Thomas A. Runkler}
\orcid{0000-0002-5465-198X}
\email{thomas.runkler@siemens.com}
\affiliation{%
  \institution{Siemens AG,}
  \institution{Technical University of Munich}
  \city{Munich}
  \state{Bavaria}
  \country{Germany}}


\begin{abstract}
  Recent advances in Tiny Machine Learning (TinyML) empower low-footprint embedded devices for real-time on-device Machine Learning (ML). While many acknowledge the potential benefits of TinyML, its practical implementation presents unique challenges. This study aims to bridge the gap between prototyping single TinyML models and developing reliable TinyML systems in production: (1)~Embedded devices operate in dynamically changing conditions. Existing TinyML solutions primarily focus on inference, with models trained offline on powerful machines and deployed as static objects. However, static models may underperform in the real world due to evolving input data distributions. We propose online learning to enable training on constrained devices, adapting local models towards the latest field conditions. (2)~Nevertheless, current on-device learning methods struggle with heterogeneous deployment conditions and the scarcity of labeled data when applied across numerous devices. We introduce federated meta-learning incorporating online learning to enhance model generalization, facilitating rapid learning. This approach ensures optimal performance among distributed devices by knowledge sharing. (3)~Moreover, TinyML's pivotal advantage is widespread adoption. Embedded devices and TinyML models prioritize extreme efficiency, leading to diverse characteristics ranging from memory and sensors to model architectures. Given their diversity and non-standardized representations, managing these resources becomes challenging as TinyML systems scale up. We present semantic management for the joint management of models and devices at scale. We demonstrate our methods through a basic regression example and then assess them in three real-world TinyML applications: handwritten character image classification, keyword audio classification, and smart building presence detection. The results confirm the effectiveness of our approaches from various perspectives, such as accuracy improvement, resource savings, and engineering effort reduction.
\end{abstract}



\received{15 December 2023}
\received[revised]{26 April 2024}
\received[accepted]{07 May 2024}

\maketitle

\section{Introduction}
\label{sec:introduction}

We distinguish between large and tiny Machine Learning~(ML). At a large scale, researchers are creating extensive ML models like "ChatGPT" on cloud clusters using numerous GPUs. These models can have carbon footprints similar to vehicles and even airplanes. The community is increasingly concerned about the expense and sustainability of deploying such large-scale models. Conversely, Tiny Machine Learning (TinyML) is a rapidly growing field bridging the gap between ML and embedded systems. It focuses on conducting sensor analytics directly on embedded devices, particularly on Internet of Things (IoT) devices, shifting data processing from the cloud to the edge. By computing data locally instead of streaming it to the cloud, TinyML offers computational efficiency, lower latency, reduced communication overhead, and improved privacy. The shift towards TinyML is also driven by the increasing prevalence of low-power and cost-effective embedded devices, known as Microcontrollers (MCUs). A recent report highlights that over 250 billion MCUs are deployed today, with continually rising demand, especially in industries\footnote{\url{https://venturebeat.com/ai/why-tinyml-is-a-giant-opportunity/}}. Envisioning a future where such tiny devices employ intelligent ML algorithms to sense their surroundings, predict events, and provide real-time recommendations is becoming realistic.

\begin{figure}[tbp]
\centering
\includegraphics[width=0.525\columnwidth]{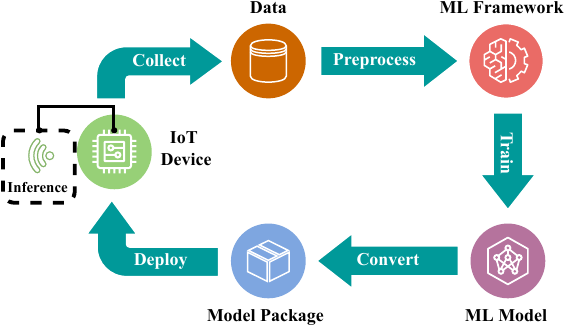}
\caption{A typical workflow for the development of a TinyML model.}
\label{fig_1}
\end{figure}

\begin{figure}[tbp]
\centering
\includegraphics[width=0.625\columnwidth]{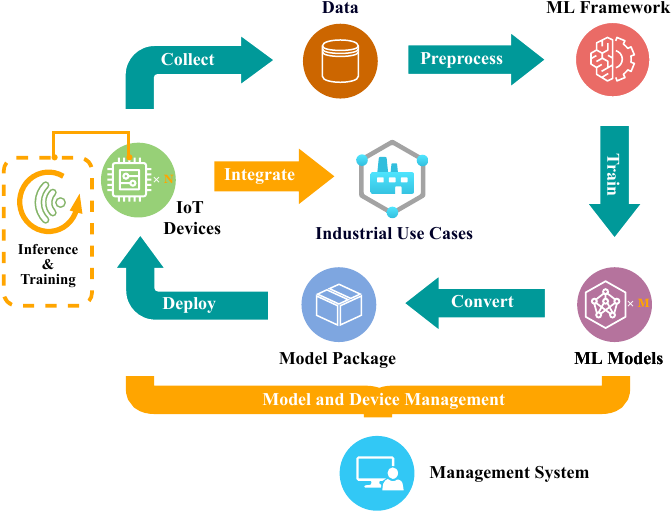}
\caption{The pipeline for developing TinyML in industrial settings. This work focuses on the previously overlooked components marked in orange.}
\label{fig_2}
\end{figure}

Figure~\ref{fig_1} illustrates a typical development workflow for TinyML models involving sensor data acquisition, preprocessing, model training, and conversion. Currently, most TinyML solutions like TensorFlow Lite Micro\footnote{\url{https://www.tensorflow.org/lite/microcontrollers}} and MicroTVM\footnote{\url{https://tvm.apache.org/docs/topic/microtvm/index.html}} follow this approach, simplifying the model creation for on-device inference. Although many organizations recognize the value of ML, merely 13\% of ML projects reach production stage\footnote{\url{https://venturebeat.com/ai/why-do-87-of-data-science-projects-never-make-it-into-production}}. We are more cautiously optimistic about TinyML, given that most embedded devices lacking software support and face more significant resource limitations, which traditional ML fails to address. Figure~\ref{fig_2} shows a workflow for applying TinyML for production. Considering the vast number of these miniature devices and models, coupled with their complex deployment conditions, effectively managing resources and ensuring models remain high-performing over time emphasize the need for robust and comprehensive strategies beyond developing single ML models.

In this work, we pinpoint three challenges in developing reliable TinyML systems in industrial settings. We then suggest corresponding approaches based on our prior research to tackle these problems. Our emphasis lies in integrating and aligning our methods, highlighting their roles from a new industrial perspective throughout the workflow depicted in Figure~\ref{fig_2}, substantiating their effectiveness through comprehensive experiments on novel applications, and providing guidance about how to leverage the combinational value of our proposed approaches:

\begin{enumerate}
    \itemsep0em
    \item Real-world conditions are subject to constant change. Following the traditional pipeline illustrated in Figure~\ref{fig_1}, we train individual TinyML models offline on powerful computers and deploy them to MCUs only for inference. However, these models are static and, once deployed, can struggle to adapt to changing input data patterns, a problem known as concept drift and data drift, resulting in performance degradation. We propose a system called TinyML with Online Learning (TinyOL)~\cite{Ren2021}, which allows incremental training on resource-constrained embedded devices. TinyOL enables single Neural Networks (NNs) to learn from streaming field data one piece at a time, keeping the model up-to-date and capable of handling "drifts" without storing historical data. 

    \item With the increasing number of TinyML clients  integrated into the system, training a universal model for all is impractical, primarily due to limited labeled data and heterogeneous conditions among distributed devices. Given the prevalence of embedded devices, it is reasonable to wonder if TinyML systems can benefit from gathering knowledge across them. We introduce TinyReptile~\cite{Ren2023a}, a model-agnostic meta-learning framework integrated with online learning, to address deployment heterogeneity. TinyReptile aggregates knowledge from devices in a federated way and collaboratively develops an NN initialization with strong generalizability. This initialization can be easily fine-tuned on new devices with very little data, ensuring optimal performance across devices. To further enhance the communication efficiency and privacy of TinyReptile, we present TinyMetaFed~\cite{Ren2023}, which incorporates techniques like partial local reconstruction and top-P\% selective communication. 

    \item As the scale and complexity of TinyML applications expand within industries, their management become essential.
    However, the TinyML ecosystem is fragmented, encompassing various hardware and models, ranging from diverse sensors and computational resources to distinct model structures and hardware requirements. This fragmentation poses challenges for effectively managing these assets. We showcase SeLoC-ML~\cite{Ren2022}, leveraging Semantic Web technologies to facilitate the joint management of TinyML resources (devices and models) on a large scale. It also enables non-experts to efficiently model, explore, and matchmake models and devices, and generate code for hardware deployment based on matching results.
\end{enumerate}

While introducing the abovementioned methods, we illustrate their functionalities through a sine wave regression example. Following that, we demonstrate these methods in streamlining three real-world TinyML applications, including two publicly available datasets: Omniglot~\cite{Lake2019} for handwritten character image classification, Speech Command~\cite{Warden2018} for keyword spotting audio classification, and one Siemens proprietary dataset for smart building presence detection. In each application, we begin with small-scale scenarios involving single TinyML devices, where we improve individual deployed models with TinyOL. Subsequently, we progress to larger-scale scenarios featuring numerous devices, recognizing that fine-tuning the model on each device with sufficient data using TinyOL can be cumbersome. Therefore, we achieve rapid adaptation across heterogeneous devices with minimum data by implementing TinyReptile and TinyMetaFed. Lastly, TinyML hold the potential for ubiquity in production. We demonstrate efficient management of diverse TinyML resources facilitated by SeLoC-ML to accelerate application development. The evaluation results confirm the value of our approaches for enhancing and maintaining TinyML in industries.

The remainder of the study is organized as follows: Section~\ref{sec:related work} reviews related work in TinyML, on-device learning, meta-learning, Semantic Web, and ML management. Section~\ref{sec:pipeline} outlines the pipeline for scaling TinyML and discusses the challenges within the system. Section~\ref{sec:approach} presents our solutions to these challenges. Section~\ref{sec:evaluation} introduces the three TinyML applications for evaluation, describes the experimental settings, and analyzes the results. Section~\ref{sec:conclusions} concludes the paper and explores future research.

\section{Related Work}
\label{sec:related work}
\subsection*{Proliferation of TinyML}

The growing momentum in TinyML is mainly demonstrated in four perspectives: algorithm, hardware, development tools, and applications. Various reviews~\cite{Han2022, Ray2022} offer a broad overview of TinyML, highlighting its challenges and opportunities. Researchers are actively developing algorithms that can use resources more efficiently without sacrificing performance. This includes techniques like model quantization~\cite{Park2023}, neural architecture search~\cite{Njor2023}, and inference optimization~\cite{Stahl2023, VanDelm2023}. Innovative hardware is emerging to boost processing in resource-frugal devices, such as in-memory computing~\cite{Houshmand2023}, neuromorphic computing~\cite{Chian2021}, and intelligent sensors~\cite{Marty2023}. Many development tools, like TensorFlow Lite for Microcontrollers~\cite{David2021} and Edge Impulse~\cite{Hymel2022}, are playing a role in streamlining the development of TinyML. Novel applications and products are transforming our lives and work, from health wearables~\cite{Zhou2023} to AR/VR~\cite{Gomez2022}. However, most TinyML solutions focus on isolated aspects of the ecosystem and lack consideration for real-world implementation. To avoid devolving the ideas into mere scientific experiments, assembling an integrated pipeline for scaling TinyML applications is crucial.

\subsection*{Advancements of On-device Learning}
Due to the limited capabilities of embedded devices, most TinyML frameworks are designed only for inference. The model training typically occurs offline on more powerful machines.
Nevertheless, there is a growing interest in exploring the feasibility of training ML models directly on devices. Work~\cite{Rajapakse2023} provides a thorough survey of updatable TinyML, introducing the concept of "reformable TinyML," which encompasses algorithms in three categories: "on-device offline learning," "online learning," and "networking reliant." In this context, we concentrate on the "online learning" approaches, wherein training is conducted directly on devices. The early endeavors in online learning primarily revolve around classical ML, wherein traditional ML algorithms are adapted for resource-constrained embedded devices, enabling on-device training. Examples include instance-based methods, such as an NN feature extractor connected with a K-Nearest-Neighbors (KNN) classifier~\cite{Disabato2020}, and linear-based methods, such as train++~\cite{Sudharsan2021}. Subsequently, with the surge of deep learning, NN designed for on-device learning have gained prominence. For instance, J. Lin et al.~\cite{Lin2022} employed sparse updates and quantization-aware scaling to train a Convolutional NN (CNN) using just 256KB of memory. However, their method is complex and is only applicable to CNN. In work~\cite{Rueb2023} and~\cite{Khan2023}, sparse backpropagation is applied to reduce the computational load of backpropagation during on-device training. TinyTL~\cite{Cai2020} has proposed freezing the model weights and training only the bias to save memory usage during learning.
Recent research has introduced forward-only training approaches~\cite{Dellaferrera2023, Pau2023}, eliminating the necessity of backpropagation during training and thus reducing the memory needed. Nevertheless, forward-only algorithms require specific updating rules to approximate backpropagation, which can cause accuracy decreases and additional computations. Our work, TinyOL~\cite{Ren2021}, stands as one of the earliest efforts to facilitate model-agnostic on-device learning by incorporating the online learning technique into NNs. By integrating online learning into existing algorithms, we process sensor data one by one and continuously update the model, leading to more efficient resource utilization and faster training.

\subsection*{Meta-Learning for "Learning How to Learn"}

Meta-learning is a promising method for rapid adaptation with limited data, making it well-suited for TinyML. Unlike transfer learning, which fine-tunes a pre-trained global model on small datasets without guaranteed performance improvements~\cite{Parnami2022}, meta-learning focuses on training a common model for easy fine-tuning. The initial approach, called "model-agnostic meta-learning" (MAML) ~\cite{Finn2017}, introduced the concept using gradient-based optimization. However, MAML requires computing higher-order derivatives, which makes it computationally intensive. Subsequent research has aimed to enhance performance~\cite{Antoniou2019} while reducing computational demands~\cite{Nichol2018}. Additionally, meta-learning has been explored in online learning~\cite{Finn2019, Acar2021} and federated learning scenarios~\cite{Fallah2020, Lin2020a}, inspiring its applicability to TinyML. Nevertheless, most existing methods are unsuitable for embedded devices. Recent efforts, including MetaLDC~\cite{Liu2023}, TinyReptile~\cite{Ren2023a}, and TinyMetaFed~\cite{Ren2023}, have emerged to apply meta-learning to constrained devices.

\subsection*{Semantic Web Technology for Information Integration}

Semantic Web technology offers a way to store and handle diverse data sources with varying structures, making it an ideal candidate for integrating heterogeneous information in TinyML. Semantics have proven valuable in industries~\cite{Rojas2021, Mehdi2017}. In the IoT realm, semantic models like Smart Sensor Network Ontology (S3N)~\cite{Sagar2018} and Sensor, Observation, Sample, and Actuator Ontology (SOSA)~\cite{Janowicz2019} are widely used to describe hardware, including their properties and interactions. The Thing Description (TD) ontology~\cite{Charpenay2016} aims to make various IoT devices accessible as web resources by using established web standards. However, IoT systems involve not only hardware but also on-device software components like ML models. Despite the efforts in hardware information modeling, very few~\cite{Dockins2016} have addressed the modeling of the software stack, especially ML models, in the context of IoT. An existing ontology~\cite{Nguyen2020} describes NN models but lacks hardware-specific information. We have proposed a semantic ontology~\cite{Ren2022b} tailored for NN models in IoT scenarios, with hardware specifications in mind, such as resource and platform requirements. Our approach bridges IoT devices and ML models with a common representation and enhances interoperability in the ecosystem.  

\subsection*{Machine Learning Management}

ML management has recently gained increased attention, particularly in TinyML. This is driven by the growing number of IoT devices and the continuous development and distribution of ML models. TensorFlow Lite Metadata\footnote{\raggedright \url{https://www.tensorflow.org/lite/convert/metadata}} and Model Card~\cite{Mitchell2019} have been introduced to describe and organize the information of trained models, increasing the transparency in distributing ML models. There is also a database~\cite{Vartak2016} designed for tracking ML models. Nevertheless, these methods face scalability issues in TinyML as they require substantial manual work and do not represent the relationships between ML models and hardware. TinyML as a service~\cite{Doyu2021} and TinyML Operations (TinyMLOps)~\cite{Antonini2022} build an abstraction layer for various hardware compilers, automatically orchestrating various ML components and generating compiled ML models for diverse hardware platforms. Although this approach supports ML model deployment across embedded platforms, it lacks features for users to store and exchange information about existing resources. Our framework~\cite{Ren2022} leverages Semantics to enable the integrated management of TinyML models and embedded devices at scale, which supports information modeling and exchange, component matchmaking, benchmarking, and code generation.

\section{Challenges in Applying TinyML in Production}
\label{sec:pipeline}

Applying TinyML in the real world and scaling its life cycle in industries introduces many challenges. Frequently, the procedures to maintain high-performing TinyML applications and manage resources are overlooked. As depicted in Figure~\ref{fig_2}, once TinyML models are deployed on hardware, they become integral parts of real-world applications. When necessary, TinyML models need to be adapted in the field to the latest environmental conditions for robust performance. Moreover, the essential advantage of embedded devices is their widespread deployment. With the increasing scale of TinyML applications, we need to maintain the high performance of all models among devices and, at the same time, address individual deployment conditions and customization requirements. More importantly, various assets (models and devices) are employed or produced throughout the process, highlighting the necessity of an efficient management system to handle these diverse components.

It is worth mentioning that embedded devices are much cheaper and more constrained than general-purpose hardware, such as laptops and phones, since they are designed to run simple tasks steadily, albeit slowly. The hardware constraints also impose limitations on software, as limited memory on embedded devices prevents the use of traditional software toolchains and requires efficient code implementations.

This work emphasizes the essential procedures for applying reliable TinyML systems in production, as highlighted in orange in Figure~\ref{fig_2}, which go beyond experimenting with single models and are equally crucial as model development. Next, we discuss three major challenges in achieving this objective.

\subsection*{Challenges}

\subsection*{Challenge \rom{1}: How to Efficiently Adapt to Changing Conditions on Single Devices?}
\label{subsubsec:challenge_1}

Models trained offline may prove ineffective in real-world scenarios because of the ever-changing deployment environments. Even data collected from the same machine at different times can exhibit substantial differences. One may apply the cyclic process depicted in Figure~\ref{fig_1} to mitigate model performance degradation. Nevertheless, transferring data from devices to a data center for model retraining is costly and susceptible to latency issues. A more practical approach is to update the models directly on devices to align with the latest conditions.
Nonetheless, embedded devices have minimal memory to store training data, and training requires additional runtime features that are not necessarily needed during inference. Furthermore, training generally demands a higher peak memory usage, potentially limiting the size and performance of deployed models. We aim to facilitate robust TinyML performance by enabling efficient on-device training with minimal resource consumption.

\subsection*{Challenge \rom{2}: How to Efficiently Adapt to Various Conditions while Achieving Optimal Performance across Heterogeneous Devices?}
\label{subsubsec:challenge_2}

TinyML is pervasive. Recent research~\cite{Grau2021} indicates that applying a global model to different devices can lead to performance drops due to non-independent and identically data distribution among devices. Moreover, these devices may have distinct objectives for their ML tasks, such as different target output classes. Simply retraining and redeploying models on each device to sustain performance can be expensive. Given the widespread use of embedded devices, it is worth exploring whether aggregating their knowledge can address this challenge. Federated learning can harness data from multiple devices to collectively train a global model. However, embedded devices work in distributed environments. A typical model trained by federated learning lacks generalizability and may not perform well on every device in real-world scenarios, as considering all possible nuances in all devices' deployment conditions can be challenging. Our goal is to achieve portability and optimal performance across devices in a federated manner while accounting for heterogeneous input data distributions and different prediction objectives.

\subsection*{Challenge \rom{3}: How to Manage Diverse TinyML Devices and Models at Scale?}
\label{subsubsec:challenge_3}
 
As development progresses, it is crucial to systematically manage all the TinyML resources (devices and models). Embedded devices, specialized for various tasks, exhibit considerable hardware diversity, including onboard sensors, computational capabilities, and runtime platforms. In the TinyML context, these devices use NNs to process sensor data and make intelligent decisions locally. To efficiently deploy TinyML models on devices, it is necessary to investigate their compatibility. NN models come in various architectures, e.g., with different layer combinations and input/output formats. Trained NN models are typically distributed as binary files without a standardized description of their structures and functionalities. All these lead to a fragmented ecosystem, hindering the efficient management and deployment of TinyML applications, especially in industrial settings involving hundreds or thousands of devices and models. It is evident that managing TinyML for production requires a deep understanding of hardware, software tools, algorithms, and applications. Our objective is to streamline the unified management of TinyML resources, facilitating the reusability and interoperability of TinyML.

\section{Approach}
\label{sec:approach}

In this section, we introduce a sine wave example as a practical illustration for the methods we will present later. Afterwards, we delve into the approaches for tackling the three challenges outlined in the last section. 
In our previous research, we have proposed various methods to address specific issues in TinyML. Now, for the first time, we are streamlining the TinyML workflow by integrating these research efforts to facilitate robust TinyML systems in production.

\subsection*{Sine Wave Regression - a Demonstration Example}
\addcontentsline{toc}{subsection}{Sine Wave Regression - a Demonstration Example}

\begin{figure}[tbp]
\centering
\includegraphics[width=0.5\columnwidth]{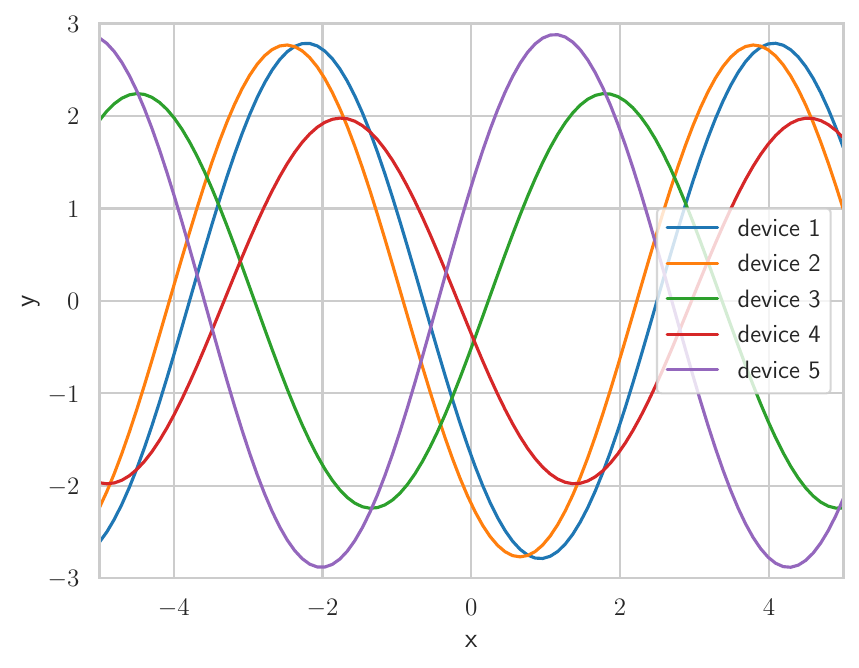}
\caption{Five random sine wave functions for five TinyML devices.}
\label{fig_3}
\end{figure}

The sine wave regression example is inspired by~\cite{Finn2017} and is defined as follows: each client/device has a task of fitting a sine function ${f(x) = a \sin(b\, x + c)}$, as shown in Figure~\ref{fig_3}. To reflect the heterogeneous data distributions among TinyML devices, we can choose the parameters ($a$, $b$, and $c$) for each sine function randomly or in a controlled manner. As we introduce our approaches later, we first explain the underlying theory of each one and demonstrate their implementations whenever possible using the sine wave example.

\subsection*{Objective \rom{1}: How to Efficiently Adapt to Changing Conditions on Single Devices?}
\addcontentsline{toc}{subsection}{Objective \rom{1}: How to Efficiently Adapt to Changing Conditions on Single Devices?}

We address the challenge of dynamically changing environments by introducing TinyML with Online Learning (TinyOL)~\cite{Ren2021}, enabling incremental training on constrained devices with streaming data. 

\subsubsection*{TinyOL}
\addcontentsline{toc}{subsubsection}{TinyOL}

\begin{figure}[tbp]
\centering
\includegraphics[width=0.65\columnwidth]{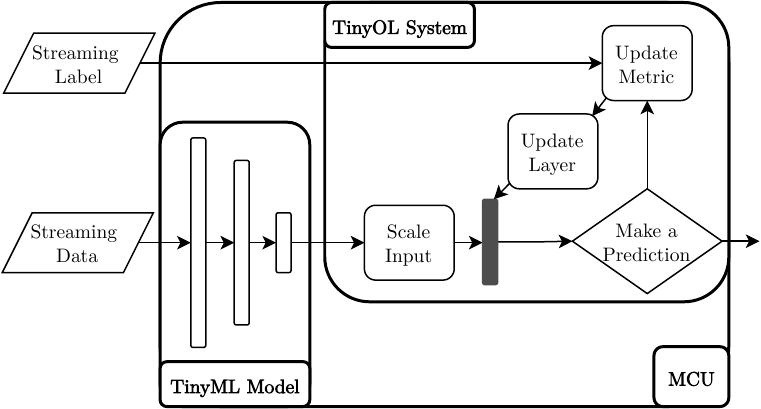}
\caption{Building blocks of TinyOL.}
\label{fig_4}
\end{figure}

TinyOL, illustrated in Figure~\ref{fig_4}, is designed for training NNs on devices with limited resources. The algorithm is based on online learning principles, which suits IoT devices with sensors that generate continuous data streams. Inference of ML models involves processing data piece by piece. TinyOL leverages this sequential property and extends existing NNs with post-training ability. Online learning allows for training models one data sample at a time, enabling adaptation to streaming data without storing a large training dataset. To our knowledge, this is one of the first methods enabling incremental on-device learning on MCUs.

The central component of TinyOL is the layer highlighted in grey in Figure~\ref{fig_4}. This layer contains a varying number of customizable neurons that can be configured, initialized, and updated at runtime, based on performance requirements and resource availability. during runtime. When connected to an existing network, this additional layer becomes the new output layer of the NN. Usually, pre-trained NNs are fixed once uploaded into the Flash memory of an MCU. However, TinyOL can train the added layer as it operates in RAM. This concept is similar to transfer learning, where we freeze part of a pre-trained model and fine-tune its last few layers.

\begin{algorithm}[tbp]
  \caption{TinyOL}
  \begin{algorithmic}[1]
    \STATE Initialize TinyML model and TinyOL system
    \FOR{x in StreamingData} 
        \STATE x\upquote \ = TinyML.Process(x)
        \STATE TinyOL.UpdateRunningMeanAndVariance(x\upquote)
        \STATE x\upquote \upquote \ = TinyOL.ScaleInput(x\upquote)
        \STATE y\upquote \ = TinyOL.Predict(x\upquote \upquote)
        \IF{y is available}
          \STATE TinyOL.UpdateMetrics(y\upquote, y) 
          \STATE TinyOL.UpdateWeights(y\upquote, y)
        \ENDIF
    \ENDFOR
  \end{algorithmic}
  \label{a_1}
\end{algorithm}

The algorithm of TinyOL is shown in Algorithm~\ref{a_1}. New data passes through the existing NN during each iteration and then into TinyOL. Depending on the task, mean and variance accumulation can be updated for input standardization. The training and prediction are performed in an interleaved fashion. After inference, if a corresponding label is available, the evaluation metrics and the weights of the additional layer can be updated using online gradient descent methods like Stochastic Gradient Descent (SGD). Once the neurons are updated, the data pair can be discarded. In other words, only one data pair resides in the memory at a time, eliminating the need to store historical data. Here, we emphasize evaluating model performance before updating weights during each iteration to ensure fair assessment.
 
Compared to traditional batch/offline training methods, TinyOL achieves model training with minimal resource consumption, allowing on-device training with massive streaming data and keeping models robust against drifts. We have developed the TinyOL framework in C/C++. The core runtime requires approximately 7 KB of RAM and 135 KB of Flash memory. This compact size makes it ideal for devices with limited resources.

\begin{figure}[tbp]
\centering
\includegraphics[width=0.565\columnwidth]{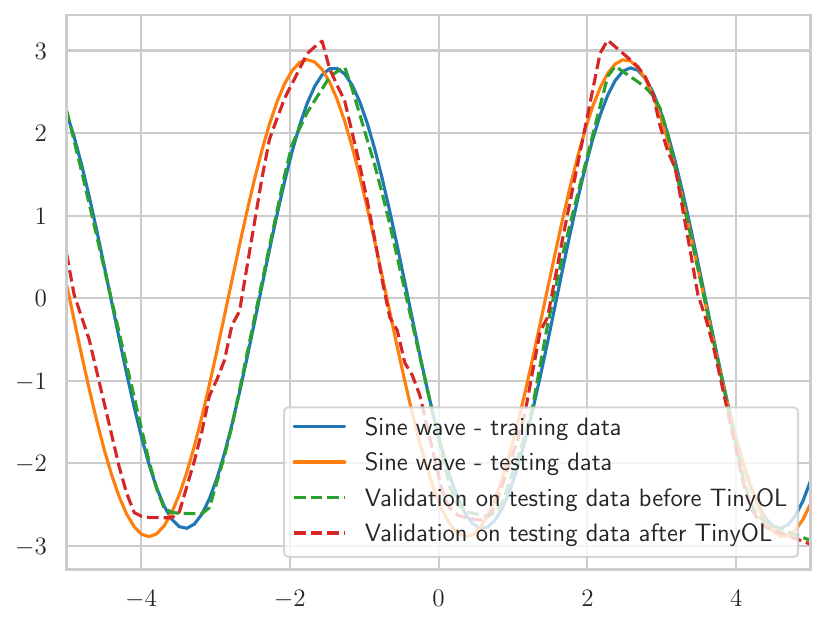}
\caption{Demonstration of TinyOL using the sine wave regression example: Suppose the data distribution between the training and testing data, labeled as "Sine wave - training data" and "Sine wave - testing data," differs due to changing real-world conditions. TinyOL helps the NN improve its performance after deployment by enabling on-device post-learning. The results on the testing data before and after TinyOL enabled are marked as "Validation on testing data before TinyOL" and "Validation on testing data after TinyOL." We can observe that after fine-tuning, the predicted output aligns much better with the distribution of the testing data.}
\label{fig_sine_tinyol}
\end{figure}

To demonstrate TinyOL on the sine wave example, we consider an IoT device with an ML task of fitting a given sine function labeled "Sine wave - training data"  in Figure~\ref{fig_sine_tinyol}. We use the sampled data points from this sine function to train an NN comprising four fully connected layers: 1 → 32 → 32 → 1. Image that after training, we deploy this trained static model to an IoT device. However, the real world is dynamic. The data in the deployment environment, referred to as "Sine wave - testing data" in Figure~\ref{fig_sine_tinyol}, exhibit statistical characteristics different from the data we used for training. When we deploy the trained model to predict the values for the new sine wave function, the results are rather undesired, marked as "Validation on testing data before TinyOL" in the figure. 

Here is where TinyOL comes into play. We replace the last layer of the trained NN with the TinyOL system and activate online training on the device. Each time, a pair of sampled data points from the new sine wave function is used to fine-tune the neurons in TinyOL. After completing the fine-tuning process, we re-evaluate the NN on "sine wave - testing data." The improved results are labeled "Validation on testing data after TinyOL" in the figure, where the predicted values closely follow the testing sine wave function. This demonstration highlights that we can adapt the NN to new working conditions and improve its performance by leveraging TinyOL.

\subsection*{Objective \rom{2}: How to Efficiently Adapt to Various Conditions while Achieving Optimal Performance across Heterogeneous Devices?}
\addcontentsline{toc}{subsection}{Objective \rom{2}: How to Efficiently Adapt to Various Conditions while Achieving Optimal Performance across Heterogeneous Devices?}

We investigate rapid adaptation of TinyML models leveraging the pervasiveness of tiny devices, emphasizing model generalizability. We first introduce TinyReptile~\cite{Ren2023a}, a framework that conducts model-agnostic meta-learning incorporating online learning in a federated context across tiny devices. To further address the computational and communication constraints of embedded devices and improve privacy, we propose TinyMetaFed~\cite{Ren2023}, which extends TinyReptile with several techniques, including partial local reconstruction, learning rate scheduling, and top P\% selective communication.

NNs generally perform better with larger training datasets. However, in many TinyML applications, data can be limited despite numerous devices handling similar tasks. With meta-learning, we train an NN that can quickly adapt or generalize to new tasks or conditions, even with limited training data. TinyReptile and TinyMetaFed focus on federated meta-learning. In federated meta-learning, we consider a group of devices, each assigned an individual ML task, denoted as $t$, from a distribution of tasks $T$. Although these tasks follow a common pattern that uses NNs with an identical structure, their classification objectives can vary. For example, in a two-class image classification scenario, one device may classify dog vs. cat while another classifies apple vs. pear. Federated meta-learning aims to find an NN initialization that can be efficiently adapted on a new device for its unique task, which is also drawn from the same distribution $T$, e.g., classifying car vs. airplane.

The training processes in TinyReptile and TinyMetaFed are iterative. They start by initializing an NN on the server, sampling a device, training the NN on the specific task $t$ on the device, adjusting the NN toward the weights trained on that device, and moving to the next device. Tasks in meta-learning are categorized into two groups: training tasks $T_{training}$ and testing tasks $T_{testing}$. Our algorithms use $T_{training}$ to learn an optimal model initialization $\phi$ that delivers good generalization performance when applied to $T_{testing}$. In summary, TinyReptile and TinyMetaFed minimize the following loss function across all the learning tasks $T$ of devices:

\begin{equation}
    \label{eq:loss_meta}
    \mathop{minimize}_{\phi} \mathbb{E}_t [ \frac{1}{2} L(\hat{\phi}^k_t, \phi^*_t)^2],
\end{equation}

where $\phi$ represents the model initialization, and $L$ denotes the distance between the optimal weights $\phi^*_t$ for the task $t$ and the fine-tuned weights $\hat{\phi}^k_t$ trained from $\phi$ for $k$ steps.

\begin{figure}[tbp]
\centering
\includegraphics[width=0.5\textwidth]{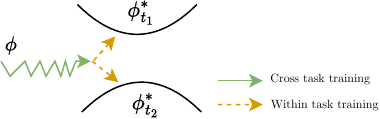}
\caption{We present a simplified weights update process in TinyReptile. Ideally, the weights $\phi$ are updated sequentially across different tasks $t$, sampled from a task distribution $T$, though we illustrate only two tasks for simplicity: $t_1$ and $t_2$. \textbf{Within each task $t$}, $\phi$ is optimized toward the optimal values $\phi^*_t$ for that task. By alternating optimizations \textbf{across tasks}, TinyReptile aims to converge to a point that minimizes the squared distance, as defined in the loss function~\ref{eq:loss_meta}.}
\label{meta_learning_illustration}
\end{figure}

\subsubsection*{TinyReptile}
\addcontentsline{toc}{subsubsection}{TinyReptile}

\begin{algorithm}[tbp]
  \caption{TinyReptile}
  \begin{algorithmic}[1]
    \STATE Central model weights $\phi$
    \STATE Server learning rate $\alpha$
    \STATE Devices, each with local streaming data $D$ and an ML task sampled from the task distribution $T$
    \STATE Device learning rate $\beta$
    \FOR{round = 1, 2, ..., i} 
        \STATE Sample one device with task $t$
        \STATE Send $\phi$ to the device
        
        \FOR{each $x$ in the local streaming data $D$} 
            \STATE Compute $\hat{\phi}^{k+1}_t = SGD(\hat{\phi}^k_t, x, \beta)$, denoting  one step of SGD on the device;
        \ENDFOR

        \STATE Send $\hat{\phi}_t$ back to the server
        \STATE Update the central model weights: 
         $\phi \leftarrow  \phi + \alpha (\hat{\phi}_t - \phi)$
    \ENDFOR
  \end{algorithmic}
  \label{a_2}
\end{algorithm}

TinyReptile is inspired by Reptile~\cite{Nichol2018}, a well-known meta-learning method recognized for its simplicity and efficiency, which achieves meta-learning by iteratively optimizing the model initialization on different tasks and progressively updating the parameters toward newly learned weights. TinyReptile extends this concept with online learning, allowing constrained devices to process local data in a streaming fashion, significantly conserving resources. TinyReptile, as described in Algorithm~\ref{a_2}, works as follows: In each round $i$, the server sends the current model weights $\phi$ to a device with an ML task $t$. The device has local training data $D$, e.g., sensor data, specific to the task t. The device then sequentially conducts several online SGD steps on $\phi$ using the streaming data $D$. Finally, the updated weights are returned to the server and integrated into the global model weights $\phi$ at a predefined learning rate $\alpha$. Essentially, the goal of TinyReptile is to find a point in the parameter space that is close to the optimal weights for all tasks, as shown in Figure~\ref{meta_learning_illustration}.

\begin{figure}[tbp]
\centering
\begin{subfigure}{0.49\textwidth}
    \centering
    \includegraphics[width=\columnwidth]{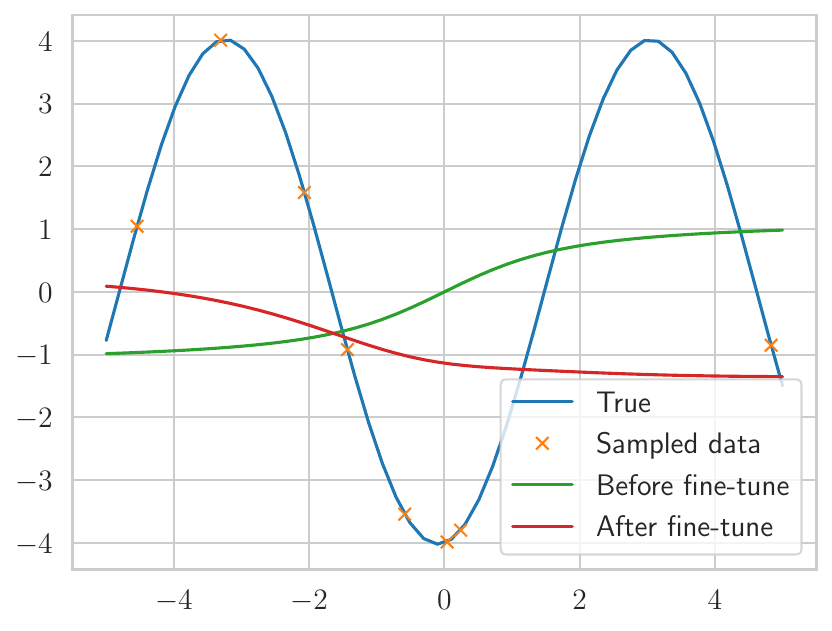}
    \caption{Federated learning (FedSGD).}
    \label{fig_sine_fedsgd}
\end{subfigure}
\begin{subfigure}{0.49\textwidth}
    \centering
    \includegraphics[width=\columnwidth]{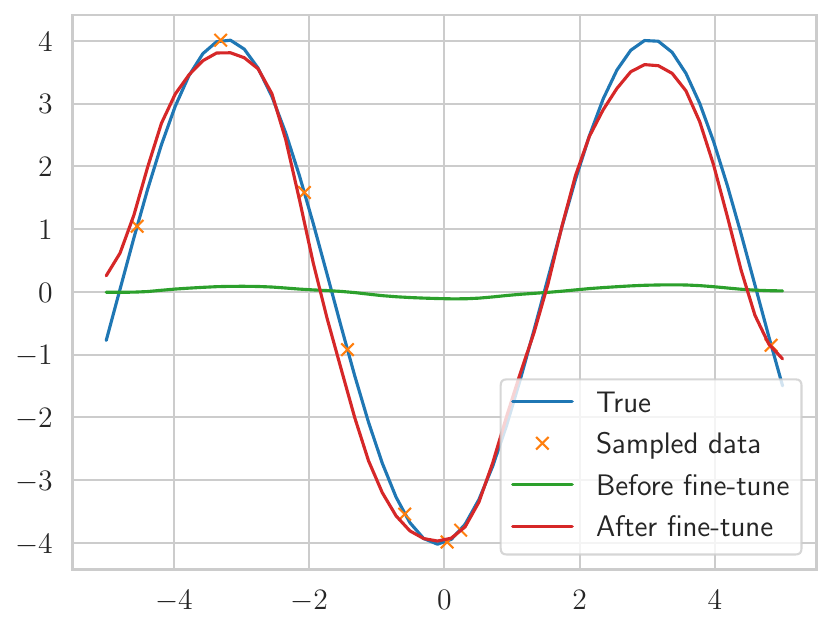}
    \caption{Meta-learning (TinyReptile).}
    \label{fig_sine_meta}
\end{subfigure}
\caption{Demonstration of federated learning (FedSGD) and meta-learning (TinyReptile) using the sine wave example: We fine-tune the NNs trained using the both methods separately by performing local SGD steps on eight sampled points from a random sine wave function. The NN comprises four fully connected layers: 1 → 32 → 32 → 1. The results indicate that the NN trained with TinyReptile rapidly converges to the sine wave and can infer values beyond the sampled points, a task that proves challenging with the NN trained with FedSGD.}
\label{fig_sine_fit}
\end{figure}

Meta-learning aims to find a model capable of quick adaptation to new tasks. Consequently, local datasets $D$ usually contain limited samples, and the training step $k$ is kept small. Our approach employs online learning, processing one data point at a time in a streaming fashion, allowing data to be discarded after each update, which is resource-efficient. Due to the serial communication schema, TinyReptile only updates one device per round,  eliminating the need for concurrent and constant connections. Any device running TinyReptile can join or leave the learning process at any time, ensuring scalability and reliability for real-world applications. Intuitively, meta-learning strives to bring the model initialization closer to a point closest to the optimal models of all the tasks, enabling rapid fine-tuning in new environments. In essence, meta-learning optimizes for generalization.

We demonstrate meta-learning through the sine wave regression example, where we compare the behaviors of federated learning (FedSGD) and meta-learning (TinyReptile), as illustrated in Figure~\ref{fig_sine_fit}. Here, each device is assigned a unique sine function ${f(x) = a \sin(b\, x + c)}$. The objective is to learn an NN initialization that can be rapidly fit to the sine function on each device with a handful of data. The outcomes show that federated learning struggles to learn a meaningful initialization in the meta-learning context. This is because federated learning aims to train an NN capable of handling all sine functions of all clients simultaneously. Consequently, a trained model ideally returns the average value of all the tasks. In the sine wave example, this translates to approximating the average values of all the sine functions $f(x)$, which are zero for all x values due to the random parameters $(a, b, c)$ in the functions.

\subsubsection*{TinyMetaFed}
\addcontentsline{toc}{subsubsection}{TinyMetaFed}

\begin{figure}[tbp]
\centering
\includegraphics[width=0.7\columnwidth]{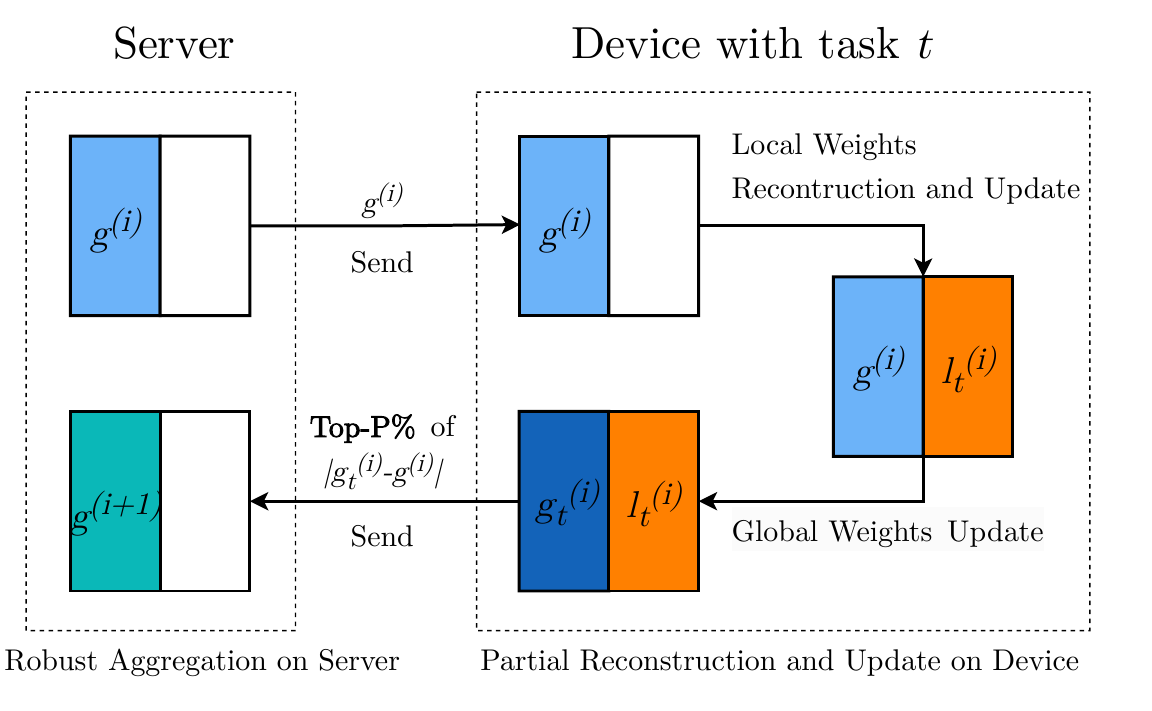}
\caption{Illustration of TinyMetaFed Workflow.}
\label{fig_6}
\end{figure}

TinyMetaFed, illustrated in Figure~\ref{fig_6}, extends TinyReptile by improving resource efficiency and client privacy through top-P\% selective communication and partial local reconstruction. Furthermore, it employs a learning rate scheduling strategy for better performance. The algorithm is outlined in Algorithm~\ref{a_3}: the weights $\phi$ of the model are divided into global weights $g$ and local weights $l$, where $\phi = \langle g, l \rangle$. The local data $D$ is separated into a support set $S$ and a query set $Q$, where $D = \langle S, Q \rangle$. In each round $i$, a device with task $t$ receives the global weights $g^{(i)}$ from the server, which it uses, along with its support set $S$, to reconstruct its local weights $l_t^{(i)}$. Afterward, the device freezes its local weights $l_t^{(i)}$ and updates the global weights $g_t^{(i)}$ using $Q$, $g^{(i)}$, and $l_t^{(i)}$. The server then receives the top-P\% updated global weights with the biggest absolute changes $|g^{(i)} - g_t^{(i)}|$ from the device instead of the entire model. Finally, the server merges these updates into the global weights $g^{(i+1)}$ using the learning rate scheduling strategy with cosine annealing:

\begin{equation}
    l(i) = \alpha_{min} + \frac{1}{2} (\alpha_{max} - \alpha_{min} - floor(\frac{i}{I}) \cdot decay) (1 + cos(\frac{\bmod(i,I)}{I} \cdot \pi)).
\end{equation}

In this formation, $i$ is the current epoch, $\alpha_{min}$ and $\alpha_{max}$ denote the learning rate range, $I$ represents the epochs between restarts, $floor(\frac{i}{I})$ tracks restart occurrences, $decay$ presents the initial learning rate's reduction, and $\bmod(i,I)$ counts epochs since the last restart. Cosine annealing starts with a high learning rate, gradually decreases to a minimum, then rapidly increases with a decay. This repeating process simulates several restarts, termed "warm restart," using previously learned weights, which can enhance training efficiency~\cite{Loshchilov2017}.

\begin{algorithm}[tbp]
  \caption{TinyMetaFed}

  \begin{algorithmic}[1]
    \REQUIRE Set of devices, each with an ML task drawn from $T$ and streaming data $D = \langle S, Q \rangle$, server learning rate scheduling function $l$.
    
    \ENSURE global weights initialization $g$

    \STATE \textbf{ServerUpdate:}
    \Indent
    \STATE Randomly initialize global weights $g$
    \FOR{each round $i$} 
    
    \STATE Sample one device with task $t$ and $D_t = \langle S_t, Q_t \rangle$
    \STATE $ g_t^{(i)} \leftarrow  $ \textbf{ClientUpdate}$(g^{(i)})$
     \STATE $g^{(i+1)} \leftarrow  g^{(i)} + l(i) (g_t^{(i)} - g^{(i)})$
    \ENDFOR
    \STATE Return $g$
    \EndIndent
    \STATE \textbf{ClientUpdate:}
    \Indent
    \STATE $D_t = \langle S_{t}, Q_{t} \rangle, g^{(i)}$
    \STATE Freeze $g^{(i)}$
    \STATE $l_t^{(i)} \leftarrow  $ \textbf{LocalWeightsReconstruction}$(S_t, g^{(i)})$
    \STATE Unfreeze $g^{(i)}$ and freeze $l_t^{(i)}$
    \STATE $g_t^{(i)} \leftarrow  $ \textbf{GlobalWeightsUpdate}$(Q_{t}, g^{(i)}, l_t^{(i)})$
    \STATE Send \textbf{top-P\%} of $g_t^{(i)}$ with the biggest changes $|g_t^{(i)}-g^{(i)}|$ back to the server
    \EndIndent
    \STATE \textbf{LocalWeightsReconstruction/GlobalWeightsUpdate:}
    \Indent
    \STATE Perform $k$ steps of SGD on the streaming data in an online learning way
    \EndIndent
  \end{algorithmic}

  \label{a_3}
\end{algorithm}

\begin{figure}[tbp]
\centering
\includegraphics[width=0.55\columnwidth]{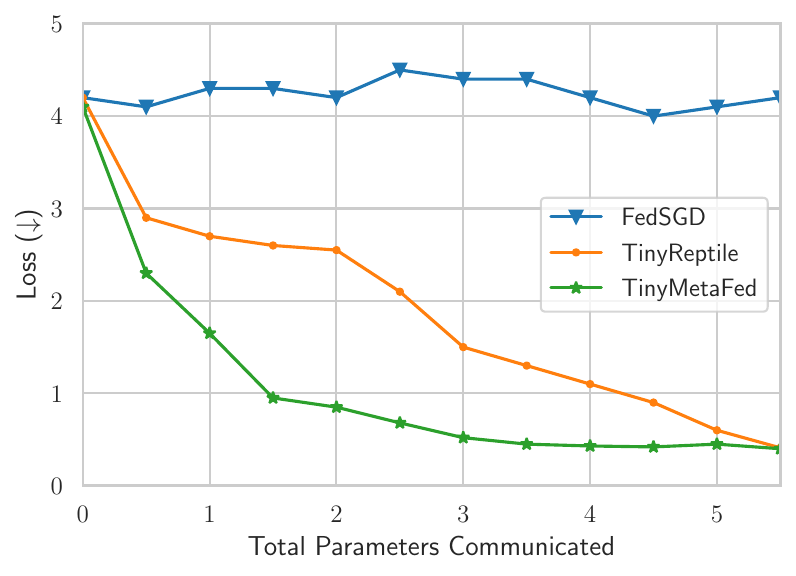}
\caption{Convergence performance of FedSGD, TinyReptile, and TinyMetaFed on the sine wave example: Loss as a function of total parameters communicated (in millions) between the server and one client. TinyMetaFed incurs lower communication and computational costs than TinyReptile, whereas FedSGD fails to converge on this task.}
\label{fig_sine_compare}
\end{figure}

We analyze the convergence performance of FedSGD, TinyReptile, and TineMetaFed, as depicted in Figure~\ref{fig_sine_compare}, which shows that TinyMetaFed achieves faster and more stable training progress than TinyReptile, whereas FedSGD cannot converge at all.

\begin{figure}[tbp]
      \centering
      \includegraphics[width=0.85\columnwidth]{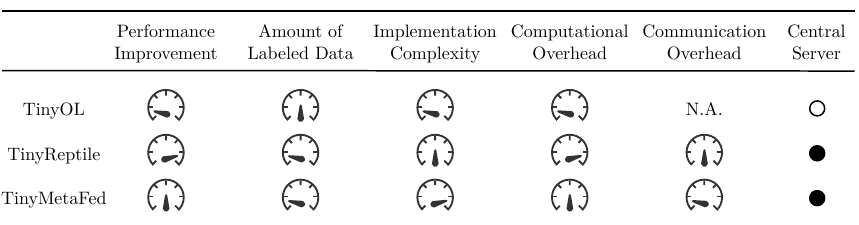}
      \caption{A qualitative comparison of TinyOL, TinyReptile, and TinyMetaFed, summarizing the experimental results detailed in Section~\ref{sec:evaluation} (The dashboard symbol represents the level of metrics from low to high; the dot symbol denotes the dependency of the infrastructure.)}
      \label{compare_diagram}
\end{figure}

Finally, we compare TinyOL, TinyReptile, and TinyMetaFed in Figure~\ref{compare_diagram} to assist readers in selecting the most suitable algorithm for on-device learning. 
Essentially, each participating device in TinyReptile conducts TinyOL except with much less training data, augmented by sharing weight updates with a central server. TinyMetaFed is an extension of TinyReptile, enriched with resource-saving and privacy-protection techniques. Consequently, the choice between pure online learning (TinyOL) and federated meta-learning (TinyReptile and TinyMetaFed) depends on the application scenarios. In cases where a central server can communicate with many devices, and each device contributes a very limited amount of labeled data (as little as 10) alongside potential individual classification objectives, opting for federated meta-learning is advisable. This decision leverages knowledge sharing across devices, enabling rapid adaptation in heterogeneous distributed conditions. Conversely, if only one or a few devices are available without a central server, but sufficient streaming training data (several dozen) is available, online learning (TinyOL) should be chosen for robust local fine-tuning on each device. Furthermore, the choice between TinyReptile and TinyMetaFed depends on the demand for resource reduction and privacy protection, as TinyMetaFed trades off between the implementation complexity and performance improvement with these two benefits.

\subsection*{Objective \rom{3}: How to Manage Diverse TinyML Devices and Models at Scale?}
\addcontentsline{toc}{subsection}{Objective \rom{3}: How to Manage Heterogeneous TinyML Devices and Models at Scale?}

We tackle the integrated management of embedded devices and TinyML models through a Semantic Web-based framework~\cite{Ren2022a}, facilitating services from describing component information and benchmarking performance to discovering potential combinations and generating engineering code. We call the framework SeLoC-ML: \textbf{Se}mantic \textbf{Lo}w-\textbf{C}ode Engineering for \textbf{ML} Applications. Acknowledging the complexity of Semantic Web techniques, we propose integrating SeLoC-ML into a low-code platform~\cite{Ren2022} to facilitate rapid design, configuration, and deployment of TinyML applications to everyone. 

\subsubsection*{SeLoC-ML}
\addcontentsline{toc}{subsubsection}{SeLoC-ML}

\begin{figure*}[t]
      \centering
      \includegraphics[width=0.975\textwidth]{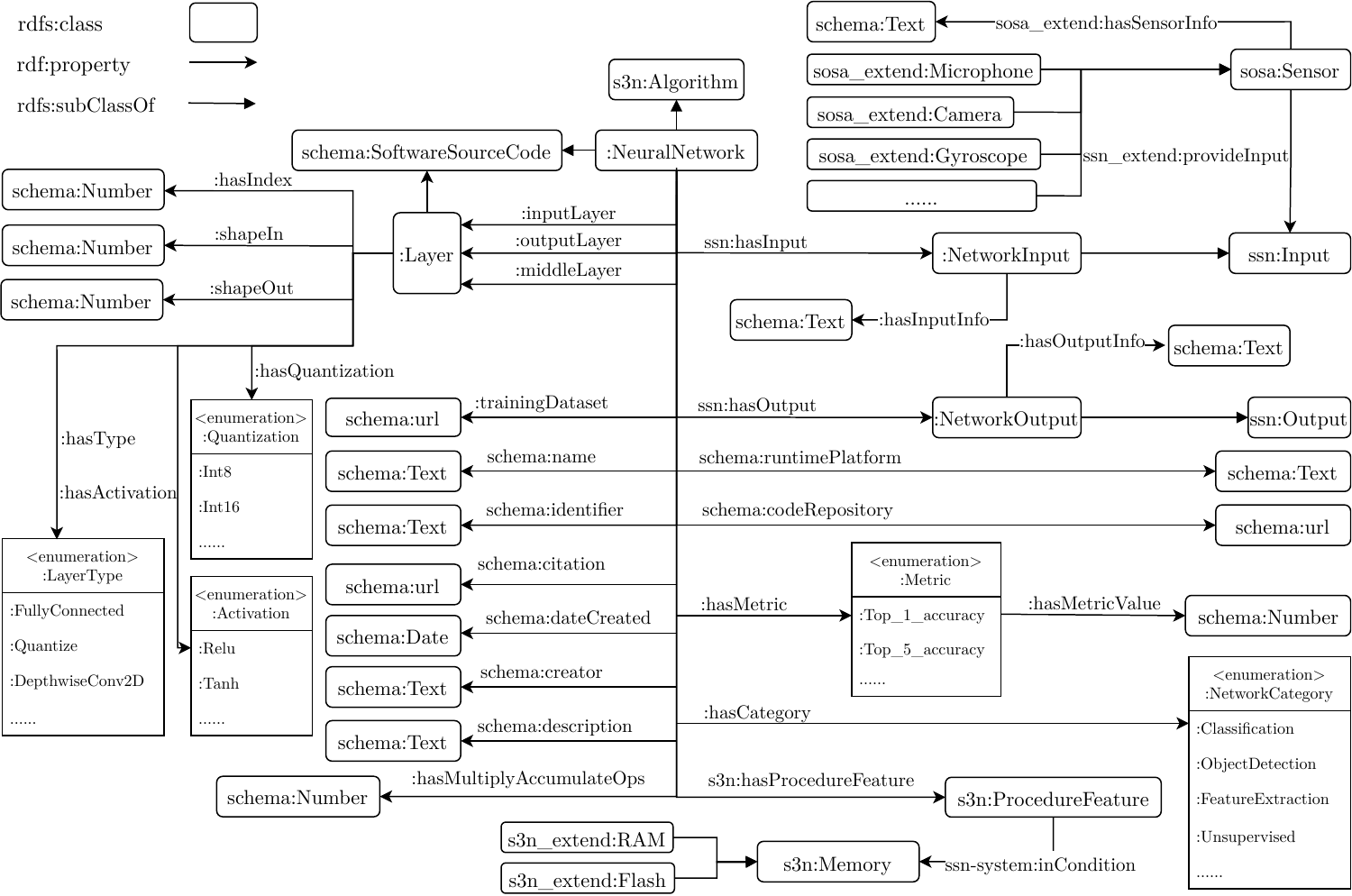}
      \caption{NN ontology for TinyML.}
      \label{nn_ontology}
\end{figure*}

In SeLoC-ML, we use two semantic models to describe heterogeneous NN models and embedded devices, respectively. While any formalized semantic models can be employed, we have chosen the standardized World Wide Web Consortium (W3C) Thing Description (TD)~\cite{Charpenay2016}  to describe IoT devices to showcase the concept in industrial settings. In alignment with TD, we have formulated a semantic ontology~\cite{Ren2022}, as shown in Figure~\ref{nn_ontology}, to describe three types of information associated with TinyML models:

\begin{enumerate}
    \item Structure: details on the NN's input/output data, architecture, and quantization.
    \item Metadata: Basic information like creation date, author, identifier, and a textual description.
    \item Hardware requirements: Information on hardware specifications, including memory and sensor demands.
\end{enumerate}

Consequently, this allows a unified representation of various knowledge about TinyML models and devices, all centrally hosted in a Knowledge Graph (KG). Graph databases can represent intricate relationships within a complex TinyML system, making vendor-agnostic knowledge management straightforward. A KG supports a range of out-of-the-box features, including knowledge discovery, similarity search\footnote{\url{https://graphdb.ontotext.com/documentation/10.2/semantic-similarity-searches.html}}, and matchmaking for TinyML models and hardware.
SPARQL\footnote{\url{https://www.w3.org/TR/rdf-sparql-query/}}, a query language, can be used for flexible interaction with a KG, making the knowledge searchable and accessible as web resources.

\begin{figure}[tbp]
      \centering
      \includegraphics[width=0.525\columnwidth]{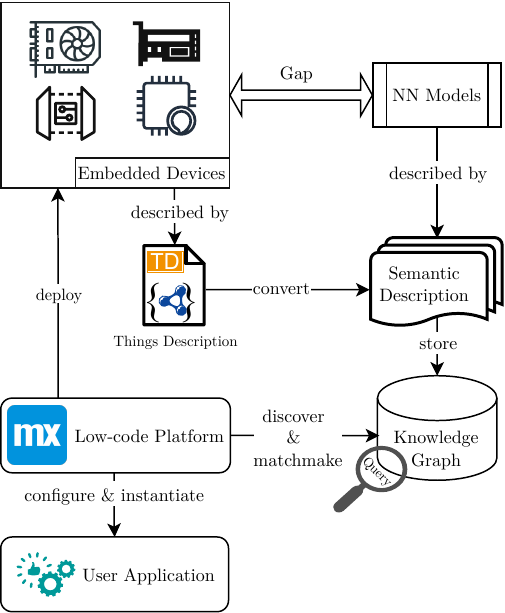}
      \caption{Framework of SeLoC-ML for managing TinyML devices and models.}
      \label{pic_7}
\end{figure}

Figure~\ref{pic_7} illustrates the system design of SeLoC-ML with two core blocks: the semantic schemas for modeling TinyML models and devices and a KG. The top of the figure highlights the challenge in TinyML, where developers face a gap between hardware (embedded devices) and software (NN models) - the absence of a unified approach to represent information in TinyML systems. We apply Semantic Web technologies to bridge this gap, transforming information about ML models and devices into semantic descriptions against the proposed ontologies and storing the knowledge in a KG. Developers can then seamlessly reuse, collaborate, and orchestrate heterogeneous information from the database to create TinyML applications in a vendor-agnostic way, enhancing reusability and interoperability within the ecosystem. Additional details on the ontologies and a demonstration of the framework in an industrial use case can be found in~\cite{Ren2022} and~\cite{Ren2022a}.

\subsubsection*{Integration into Mendix - A Low-code Engineering Platform}
\addcontentsline{toc}{subsubsection}{Integration into Mendix - A Low-code Engineering Platform}

\begin{figure}[tbp]
      \centering
      \includegraphics[width=0.95\columnwidth]{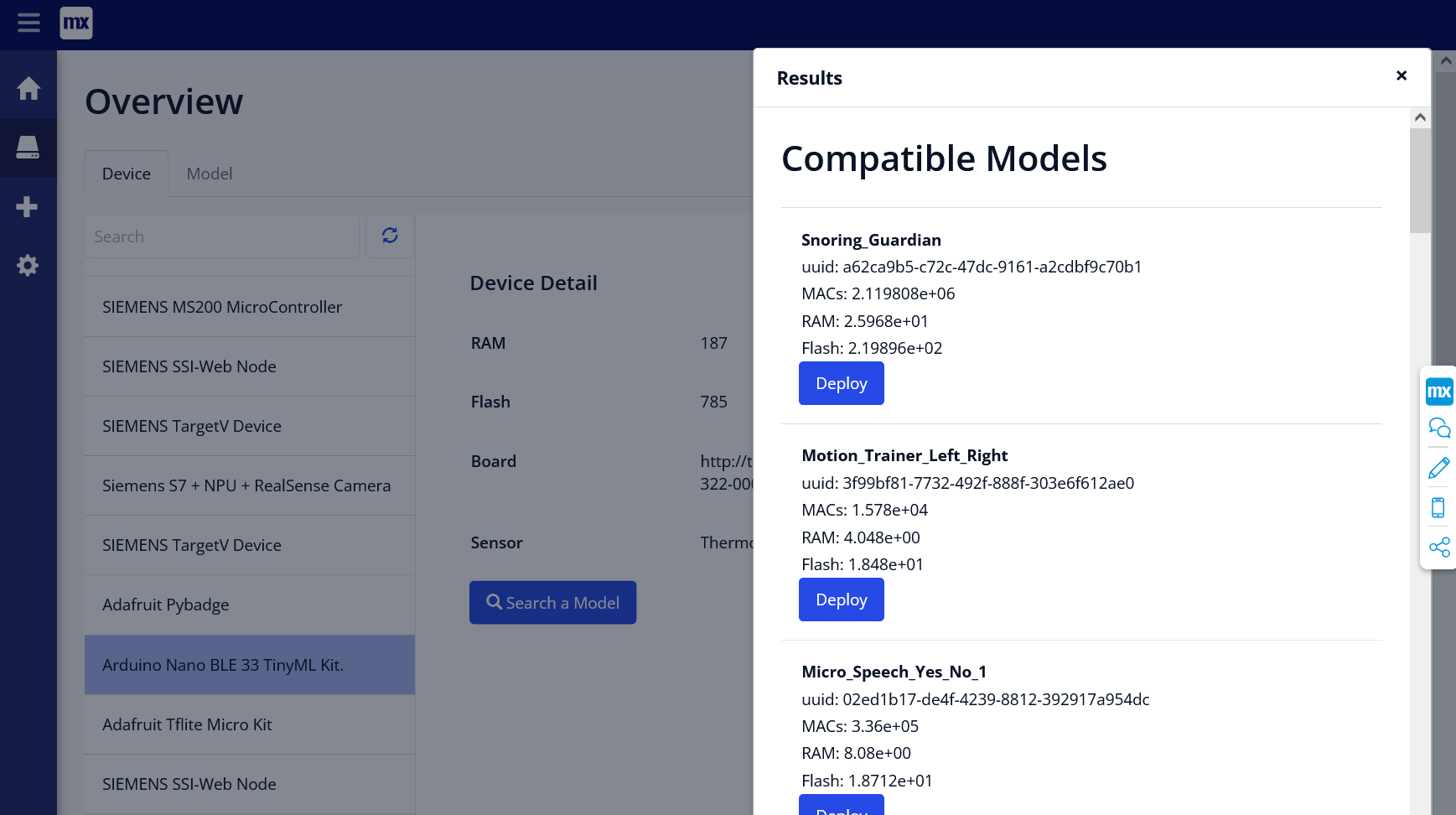}
      \caption{Integration of SeLoC-ML in the Siemens low-code platform Mendix.}
      \label{pic_8}
\end{figure}

Recognizing that not all TinyML engineers are versed in semantics, especially in SPARQL query creation, we show how TinyML management might look in the future by leveraging a low-code platform. Low-code software development minimizes the need for coding by utilizing visual interfaces in a drag-and-drop manner. This concept enables fast creation, effortless maintenance, and easy usage of enterprise-level applications without extensive programming. While various low-code platforms are available, we have opted for the Siemens low-code platform - Mendix\footnote{\url{https://www.mendix.com}} for our convenience. As depicted in Figure~\ref{pic_7}, we expand the workflow to incorporate Mendix, abstracting background engineering processes and enabling non-experts to utilize semantic services. A snapshot of the platform interface is shown in Figure~\ref{pic_8}, where we search compatible NN models for a selected IoT device. In the background, Mendix automatically generates queries based on user inputs and retrieves the desired answers from the KG. Upon matchmaking, different deployment options become available. An engineering project, ready for deployment, can be generated based on user configurations and the retrieved information, streamlining deployment on diverse hardware platforms.

In our GitHub repository\footnote{\url{https://github.com/Haoyu-R/How-to-Manage-TinyML-at-Scale}}, code and examples are available for automatically generating a semantic representation for any NN model in the ".tflite" format to be stored in a KG. This format is supported by TensorFlow Lite for Microcontrollers. The NN model used in the sine wave example, along with other models, is converted and saved in a ".ttl" turtle file. Interested users can find the turtle file and the Mendix application package in the repository.

\section{Real-world Applications}
\label{sec:evaluation}

In this section, we assess the approaches from the previous section for three practical applications: handwritten character image classification using the Omniglot Dataset, keyword spotting audio classification using the Speech Command Dataset, and smart building presence detection using a proprietary Siemens dataset.
In each application, we begin by using TinyOL to enable local on-device online learning, mitigating performance degradation on single IoT nodes.

Subsequently, we demonstrate how TinyReptile and TinyMetaFed support federated meta-learning, enhancing model generalizability for fast and customized deployment across distributed devices. Finally, we demonstrate in two case studies that SeLoC-ML can substantially reduce engineering effort in constructing and managing TinyML applications compared to the traditional approach.

\subsection*{Experimental Setup}
\addcontentsline{toc}{subsection}{Experimental Setup}

We conduct experiments using a Raspberry Pi 4 Model B\footnote{\url{https://www.raspberrypi.com/products/raspberry-pi-4-model-b/}} for the image and audio classification applications and an Arduino Nano BLE 33 MCU\footnote{\url{https://docs.arduino.cc/hardware/nano-33-ble-sense}} for the presence detection application. Table~\ref{tab:model_overview} introduces the models used in the three applications and their RAM and flash memory requirements. We construct the CNNs in the first two applications referring to the models defined in the MLPerf Tiny benchmark~\cite{Banbury2021}, primarily utilizing the "depthwise convolution" structure. In contrast, the NN used in the last application comprises several fully connected layers.
The model training is conducted using gradient descent optimizers, such as SGD and ADAM.

\begin{table}[tbp]
\centering
\caption{Overview of the models used in the applications.}
\label{tab:model_overview}
\begin{tabular}{lllll}
\hline
Application & Model Type & RAM & Flash & Parameters \\ \hline
\begin{tabular}[c]{@{}l@{}}Handwritten Character\\ Image Classification\end{tabular} &
  \begin{tabular}[c]{@{}l@{}}Convolutional\\ (5-classes)\end{tabular} &
  79.1 KB &
  475.7 KB &
  108229 \\
\begin{tabular}[c]{@{}l@{}}Keyword Spotting\\ Audio Classification\end{tabular} &
  \begin{tabular}[c]{@{}l@{}}Convolutional\\ (4-classes)\end{tabular} &
  347.4 KB &
  485.8 KB &
  105124 \\ 
\begin{tabular}[c]{@{}l@{}}Smart Building\\ Presence Detection\end{tabular} &
  \begin{tabular}[c]{@{}l@{}}Fully Connected\\ (Binary)\end{tabular} &
  3.5 KB &
  47.8 KB &
  4513 \\ \hline
\end{tabular}%
\end{table}

The evaluation metrics for TinyOL cover inference time, training time, energy consumption, performance improvement, and memory overhead. We compare TinyOL with an NN-based feature extractor + KNN classifier approach~\cite{Disabato2020}, which has a similar design to TinyOL. In the comparison, the same base NN model is connected to TinyOL layers and a KNN classifier, respectively. In the first application, we use 250 samples during the fine-tuning phase for the KNN approach, whereas in the second and third applications, we use datasets with 500 samples for fitting the KNN classifier. Since KNN is an instance-based algorithm and we choose "brute force" as the search algorithm to compute the nearest neighbors, we do not consider training overhead in this context, as it only requires storing the training data in memory. In comparing TinyReptile and TinyMetaFed with Reptile, we consider time consumption, communication cost, energy consumption, memory requirement, and prediction performance. We use a batch size of eight in Reptile, and TinyReptile serves as the baseline with a value of one for evaluating communication costs, communicating the entire model in each round. We assess energy consumption using a USB multimeter by subtracting idle energy usage from the total energy consumed during algorithm execution. The results are measured on a single device for one round, starting with the central server sending the model to a local device and ending with the server receiving the updated weights. Where possible, we repeat the experiments multiple times, and the results are presented as the mean value with standard deviation.

Throughout the evaluation, we experimented with various hyperparameters for the algorithms that can exhibit good performance, such as the learning rate (ranging from 0.001 to 0.02), though we did not fine-tune them for optimal results. Nevertheless, we keep consistent settings for these factors within each application. We exclude other algorithms in our experiments, as they proved unsuitable for TinyML, require specialized hardware like FPGA, or are dependent on the application.

\subsection*{Application \rom{1}: Handwritten Character Image Classification}
\addcontentsline{toc}{subsection}{Application \rom{1}: Handwritten Character Image Classification}

\begin{figure*}[tb]
     \centering
    
     \begin{subfigure}[b]{0.32\textwidth}
         \centering
         \includegraphics[width=\textwidth]{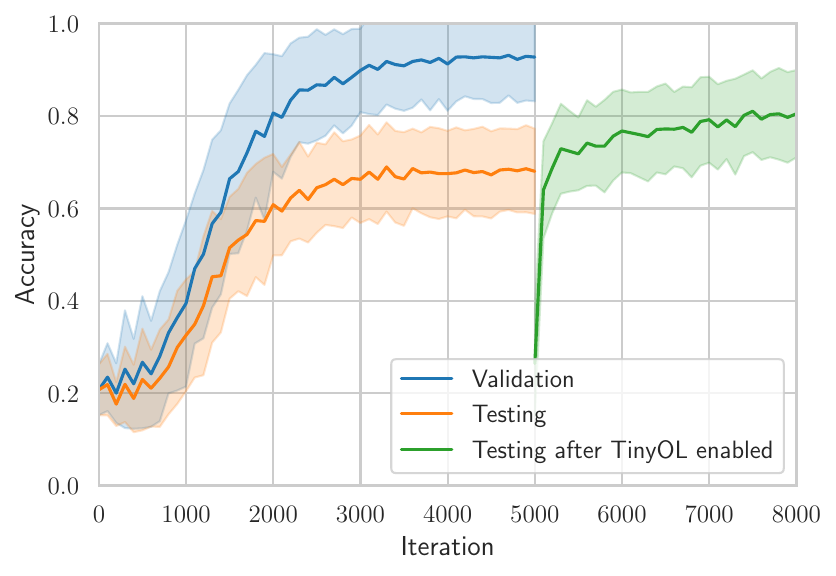}
         \caption{Handwritten Character Image Classification (5-classes).}
         \label{fig_tinyol_omniglot}
     \end{subfigure}
     \hfill
     \begin{subfigure}[b]{0.32 \textwidth}
         \centering
         \includegraphics[width=\textwidth]{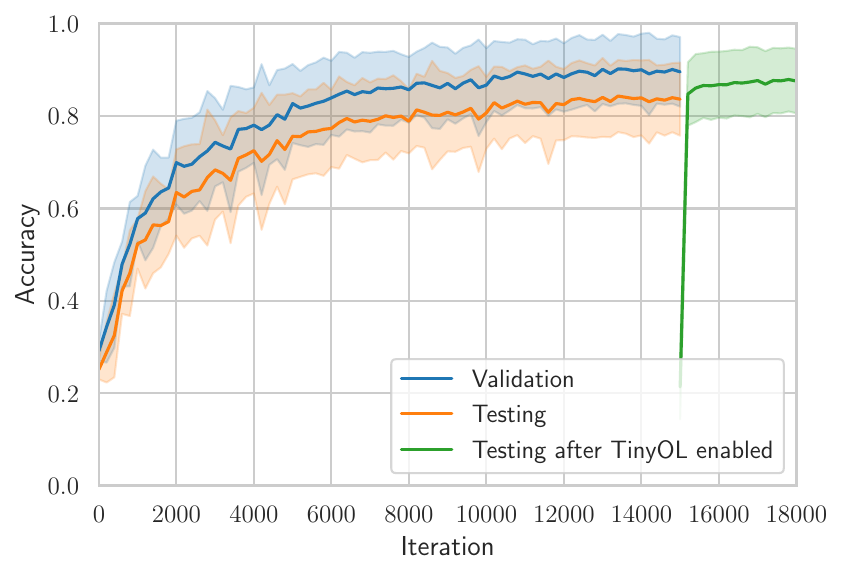}
         \caption{Keyword Spotting Audio Classification (4-classes).}
         \label{fig_tinyol_keywords}
     \end{subfigure}
     \hfill
     \begin{subfigure}[b]{0.32\textwidth}
         \centering
         \includegraphics[width=\textwidth]{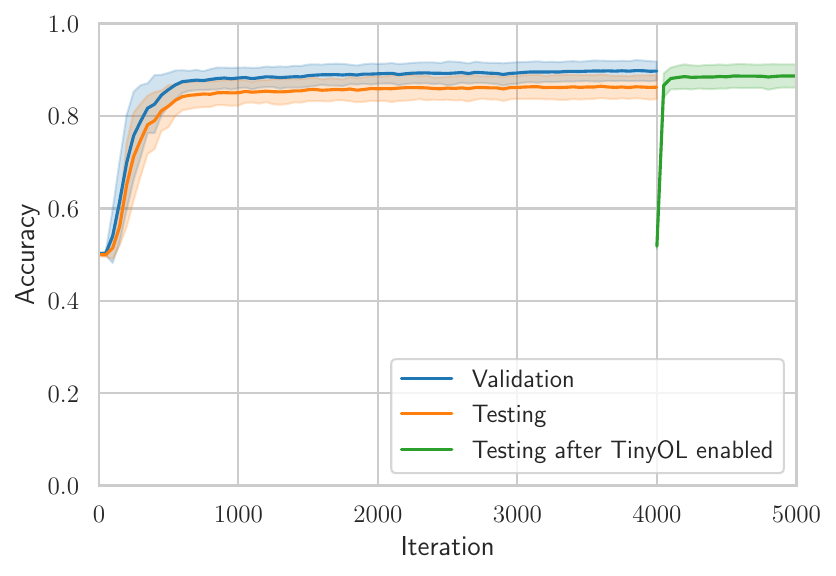}
         \caption{Smart building Presence Detection (Binary).}
         \label{fig_tinyol_smart_building}
     \end{subfigure}
        \caption{Illustration of TinyOL in the three applications. The convergence performance of the models on the training data, marked as "Validation," degrades when tested on the testing data, denoted as "Testing," which has different data distributions. The green lines in the figures show that we can improve model performance after model deployment by enabling TinyOL.}
        \label{fig_tinyol}
\end{figure*}

\begin{figure*}[tbp]
     \centering
     \begin{subfigure}[b]{0.32\textwidth}
         \centering
         \includegraphics[width=\textwidth]{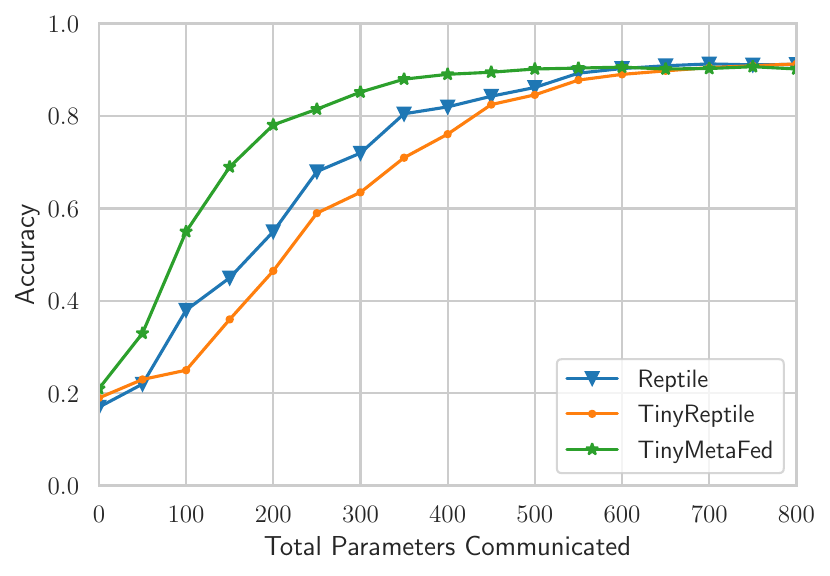}
         \caption{Handwritten Character Image Classification (5-classes).}
         \label{fig_meta_omniglot}
     \end{subfigure}
     \hfill
     \begin{subfigure}[b]{0.32\textwidth}
         \centering
         \includegraphics[width=\textwidth]{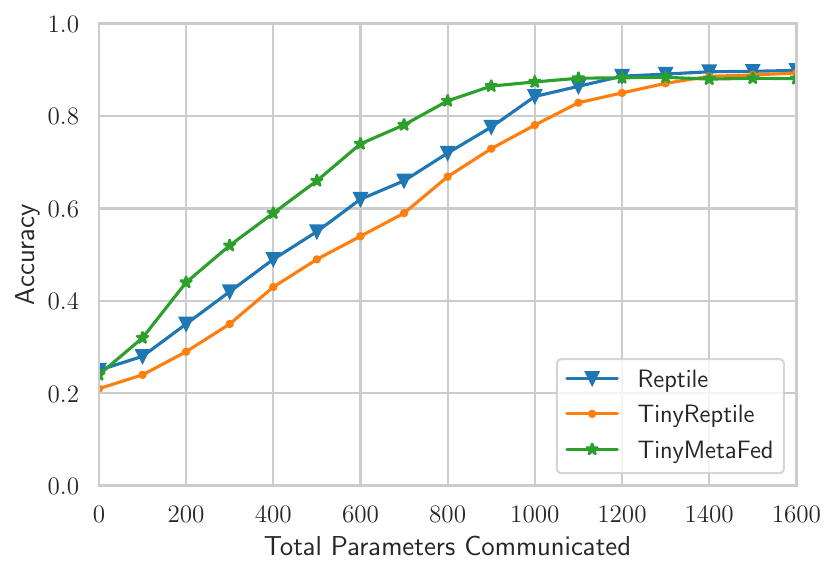}
         \caption{Keyword Spotting Audio Classification (4-classes).}
         \label{fig_meta_keyworkds}
     \end{subfigure}
     \hfill
     \begin{subfigure}[b]{0.32\textwidth}
         \centering
         \includegraphics[width=\textwidth]{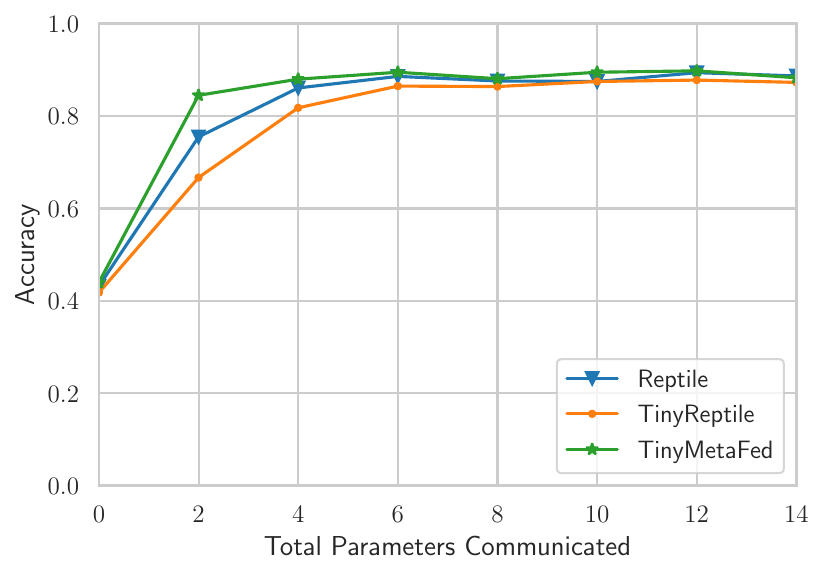}
         \caption{Smart Building Presence Detection (Binary).}
         \label{fig_meta_smart_building}
     \end{subfigure}
        \caption{Accuracy against the total parameters communicated (in millions) between the server and a single client in the three applications. We demonstrate that TinyReptile and TinyMetaFed can achieve comparable results to Reptile. Remarkably, TinyMetaFed substantially minimizes communication costs, thereby reducing overall energy consumption.}
        \label{fig_meta}
\end{figure*}

In our first application, we explore an image classification application using the Omniglot dataset. The Omniglot image dataset comprises 1623 characters representing 50 alphabets, with up to 60 image samples available for each character. 

\subsubsection*{TinyOL}

Consider a scenario where we have trained a five-class image classification model using images of five randomly selected characters from the dataset. However, upon deploying the model, we noticed that the characters presented in the real-world images are slightly positioned differently, such as rotated, shifted, or zoomed. Here, we simulate the real-world heterogeneity by applying random image augmentation to the testing data. These include rotation up to 10 degrees, translations up to 10\% in the left, right, up, and down directions, and zoom effects up to 10\% in height and width dimensions. As a result, the model performance is decreased by up to 20\%, as shown in Figure~\ref{fig_tinyol_omniglot}. This happens because the training and testing data have different statistical characteristics. To mitigate the issue, we replace the trained model's last layer with the TinyOL system and perform on-device post-training in the field (on the testing data). Although the original model remains frozen after deployment to save resources, TinyOL can fine-tune the attached new layer, resulting in a 12.4\% accuracy improvement. 

We compare TinyOL with the feature extractor + KNN classifier approach~\cite{Disabato2020}, where the original frozen NN model serves as the feature extractor and is connected to a KNN classifier. Table~\ref{tab:tinyoL_benchmark} presents the benchmark results conducted on a Raspberry Pi 4. Notably, the inference time and energy consumption with TinyOL are lower than the corresponding values with the KNN method. For instance, the inference time for one iteration is 0.048~s with TinyOL, compared to 0.054~s with the KNN method. Additionally, TinyOL has an advantage in memory overhead, with 808~KB compared to 2704~KB for batch training and 839~KB with the KNN method. Nevertheless, the KNN approach has the advantage during the training phase, requiring only the storage of instances in memory, resulting in minimal training overhead, provided there's sufficient memory space. This raises concerns with instance-based methods like KNN since additional training steps (i.e., storing more training data) lead to increased memory requirements, unlike TinyOL, where memory consumption remains constant. These results demonstrate the efficiency of the TinyOL system and its potential benefits in reducing resource consumption compared to traditional approaches.


\begin{table}[tbp]
\centering
\caption{Benchmarking TinyOL with the feature extractor + KNN approach for the three applications: We conduct experiments in Applications \rom{1} and \rom{2} using a Raspberry Pi 4 and in Application \rom{3} using an Arduino Nano BLE 33. The results for TinyOL are displayed on the \textbf{left} side of each cell, while the results for the feature extractor + KNN approach are on the \textbf{right} side. For memory overhead, we also extend the comparison with training the NN using a batch size of 128. We measure the time and energy consumption for a single iteration. It is noteworthy that the feature extractor + KNN approach, where training involves only storing relevant instances in memory, has negligible training time and energy consumption.
}
\label{tab:tinyoL_benchmark}
\resizebox{\textwidth}{!}{%
\begin{tabular}{@{}llllllll@{}}
\toprule
 &
  \begin{tabular}[c]{@{}l@{}}Inference\\ Time\end{tabular} &
  \begin{tabular}[c]{@{}l@{}}Training\\ Time\end{tabular} &
  \begin{tabular}[c]{@{}l@{}}Inference Energy\\ Consumption\end{tabular} &
  \begin{tabular}[c]{@{}l@{}}Training Energy\\ Consumption\end{tabular} &
  \begin{tabular}[c]{@{}l@{}}Final\\ Accuracy\end{tabular} &
  \begin{tabular}[c]{@{}l@{}}Accuracy \\ Improvement\end{tabular} &
  \begin{tabular}[c]{@{}l@{}}Memory Overhead\\ (TinyOL vs. Batch Training \\ vs. Feature Extractor + KNN)\end{tabular} \\ \midrule
\begin{tabular}[c]{@{}l@{}}Handwritten Character\\ Image Classification\end{tabular} &
  \textbf{0.048 s} / 0.054 s &
  0.023 s / - &
  \textbf{0.068 J} / 0.078 J &
  0.032 J / - &
  \textbf{80.6\%}  / 78.4\% &
  \textbf{12.4\%} / 10.2\% &
  \textbf{808 KB} / 2704 KB / 839 KB \\
\begin{tabular}[c]{@{}l@{}}Keyword Spotting\\ Audio Classification\end{tabular} &
  \textbf{0.044 s} / 0.046 s &
  0.019 s / - &
  \textbf{0.063 J} / 0.067 J &
  0.027 J / - &
  \textbf{86.7\%} / 85.1\% &
  \textbf{4.4\%} / 2.8\% &
  \textbf{439 KB} / 4598 KB / 503 KB \\
\begin{tabular}[c]{@{}l@{}}Smart Building\\ Presence Detection\end{tabular} &
  \textbf{0.018 s} / 0.032 s &
  0.023 s / - &
  \textbf{0.0044 J} / 0.0076 J &
  0.0056 J / - &
  88.4\% / \textbf{89.8\%} &
  2.2\% / \textbf{3.6\%} &
  \textbf{19 KB} / 100 KB / 83 KB \\ \bottomrule
\end{tabular}%
}
\end{table}

\begin{table}[tbp]
\centering
\caption{Handwritten Character Image Classification: Benchmark of the meta-learning methods on a Raspberry Pi 4. The results are measured on one device for one round.}
\label{tab:omniglot}
\begin{tabular}{@{}llllllll@{}}
\toprule
 &
  Receiving &
  \begin{tabular}[c]{@{}l@{}}Local\\ Training\end{tabular} &
  Sending &
  Total &
  \begin{tabular}[c]{@{}l@{}}Communication\\ Cost\end{tabular} &
  \begin{tabular}[c]{@{}l@{}}Memory\\ Requirement\end{tabular} &
  \begin{tabular}[c]{@{}l@{}}Energy\\ Consumption\end{tabular} \\ \midrule
Reptile     & 3.7 s & 11.3 s & 2.8 s & 17.8 s & 1 * N & 2096 KB & 37 J \\
TinyReptile & 3.7 s & 5.4 s & 2.8 s & \textbf{11.9 s} & \textbf{1}     & \textbf{641 KB}  & \textbf{24 J} \\
TinyMetaFed & 1.3 s & 5.2 s  & 0.7 s & \textbf{7.2 s}  & \textbf{0.31}  & \textbf{641 KB}  & \textbf{13 J} \\ \bottomrule
\end{tabular}%
\end{table}

\subsubsection*{TinyReptile \& TinyMetaFed}

In practical deployments involving many devices, individual devices may have unique ML task requirements, such as classifying different characters. Additionally, they may find themselves in varying deployment environments. Meta-learning enables each device to quickly adapt to its specific conditions with very little data. Here, we examine a group of devices, assigning each device a classification task involving a subset of the characters ($M=5$ classes). Although all the devices share the same number of classes, the classification characters are chosen randomly for each device. Figure~\ref{fig_meta_omniglot} compares the training progress of Reptile, TinyReptile, and TinyMetaFed based on the total number of communicated parameters. Table~\ref{tab:omniglot} presents the hardware benchmarking results on a Raspberry Pi 4.

We observe that TinyReptile and TinyMetaFed can achieve comparable final performance to Reptile, while TinyMetaFed stands out regarding communication cost and training speed, exhibiting approximately 33\% and 42\% less communication overhead compared to Reptile and TinyReptile, respectively. Reptile and TinyReptile have identical communication time consumption since both transmit the entire model between the server and a client in each round. In contrast, TinyMetaFed only transfers part of the model, significantly reducing communication overhead. Due to the batched communication schema, Reptile can require concurrent communication with N devices in each round. Notably, TinyReptile and TinyMetaFed achieve at least a twofold reduction in the local training time and up to a threefold decrease in memory requirement compared to Reptile. This improvement is attributed to online learning, which allows TinyReptile and TinyMetaFed to process local data sequentially. In comparison, Reptile needs to store historical data locally and process them in batches. The combination of reduced communication overhead, lower local resource utilization, and faster training process contribute to the lower energy consumption of TinyReptile and TinyMetaFed, with a 35\% and 65\% reduction compared to Reptile, respectively. Furthermore, TinyMetaFed enhances privacy protection by design. Many previous attack methods may become ineffective since TinyMetaFed only communicates part of the global model parameters to the server in each round, which are calculated using a subset of local data.

\begin{table}[tbp]
\centering
\caption{Testing accuracy of TinyReptile as a function of dataset size used for local fine-tuning.}
\label{tab:tinyreptile_support_set}
\begin{tabular}{@{}lllllll@{}}
\toprule
 & $D=0$ & $D=1$ & $D=2$ & $D=4$ & $D=8$ & $D=16$ \\ \midrule
\begin{tabular}[c]{@{}l@{}}Handwritten Character \\ Image Classification\end{tabular} & 20.6 & 58.3 & 75.0 & 85.9 & 91.4 & 92.1 \\
\begin{tabular}[c]{@{}l@{}}Keyword Spotting\\ Audio Classification\end{tabular}       & 24.6 & 61.7 & 76.7 & 85.1 & 89.4 & 90.4 \\
\begin{tabular}[c]{@{}l@{}}Smart Building\\ Presence Detection\end{tabular}           & 48.3 & 68.3 & 79.6 & 86.1 & 88.6 & 88.3 \\ \bottomrule
\end{tabular}
\end{table}

\begin{table}[tbp]
\centering
\caption{The analysis of various strategies applied within TinyMetaFed: Top-P\% selective communication, learning rate scheduling with cosine annealing, and partial local reconstruction.
We show the accuracy (on the left) and communication cost per round (on the right) for each setting.}
\label{tab:tinymetafed_hyperparameter}

\begin{tabular}{@{}llllll@{}}
\toprule
&
  \begin{tabular}[c]{@{}l@{}}P=100\%,\\ Without learning \\ rate scheduling, \\ Without partial \\ local reconstruction\\ (TinyReptile)\end{tabular} &
  \begin{tabular}[c]{@{}l@{}}P=100\%, \\ Without learning \\ rate scheduling\end{tabular} &
  P=100\% &
  P=75\% &
  P=50\% \\ \midrule
\begin{tabular}[c]{@{}l@{}}Handwritten Character\\ Image Classification\end{tabular} & 91.4\% / 1 & 90.0\% / 0.35 & 90.6\% / 0.35 & 89.4\% / 0.31 & 87.3\% / 0.26 \\
\begin{tabular}[c]{@{}l@{}}Keyword Spotting\\ Audio Classification\end{tabular}      & 89.4\% / 1 & 88.1\% / 0.42 & 88.3\% / 0.42 & 87.2\% / 0.36 & 85.9\% / 0.29 \\
\begin{tabular}[c]{@{}l@{}}Smart Building\\ Presence Detection\end{tabular}          & 88.6\% / 1 & 89.3\% / 0.53 & 89.6\% / 0.53 & 88.7\% / 0.46 & 86.5\% / 0.39 \\ \bottomrule
\end{tabular}
\end{table}

Subsequently, we study the parameters in TinyReptile and TinyMetaFed and offer guidance to the reader for real-world deployment. Firstly, we investigate the impact of the local dataset size $D$ on testing clients in TinyReptile, considering scenarios where TinyML devices have very limited local data for adaptation. Results in Table~\ref{tab:tinyreptile_support_set} show that the trained initialization poorly generalizes without any data. However, providing just one data pair for adaptation boosts performance by 37.7\%, with accuracy saturating at 92.1\% as $D$ increases. This suggests that a small amount of labeled data is sufficient for local fine-tuning. Further improvement may be possible with more data samples since fine-tuning is conducted in a gradient-descent way, although the effect is less noticeable. 

Next, we explore the strategies introduced in TinyMetaFed, outlined in Table~\ref{tab:tinymetafed_hyperparameter}. The first configuration strips away all proposed strategies, essentially representing TinyReptile. Significant communication savings are achieved through partial local reconstruction in the second configuration, with a communication cost of 0.35 and a similar performance of 90.6\%. In the third configuration, introducing learning rate scheduling mitigates performance decline slightly by 0.6\%. Moreover, a smaller parameter value $P$ can reduce communication costs but may compromise performance. For instance, setting P=50\% yields 87.3\% accuracy with a cost of 0.26. This suggests that weights with larger absolute values play a crucial role in model training, containing more valuable information.

\subsubsection*{SeLoC-ML}

\begin{figure}[tbp]
      \centering
      \includegraphics[width=0.75\columnwidth]{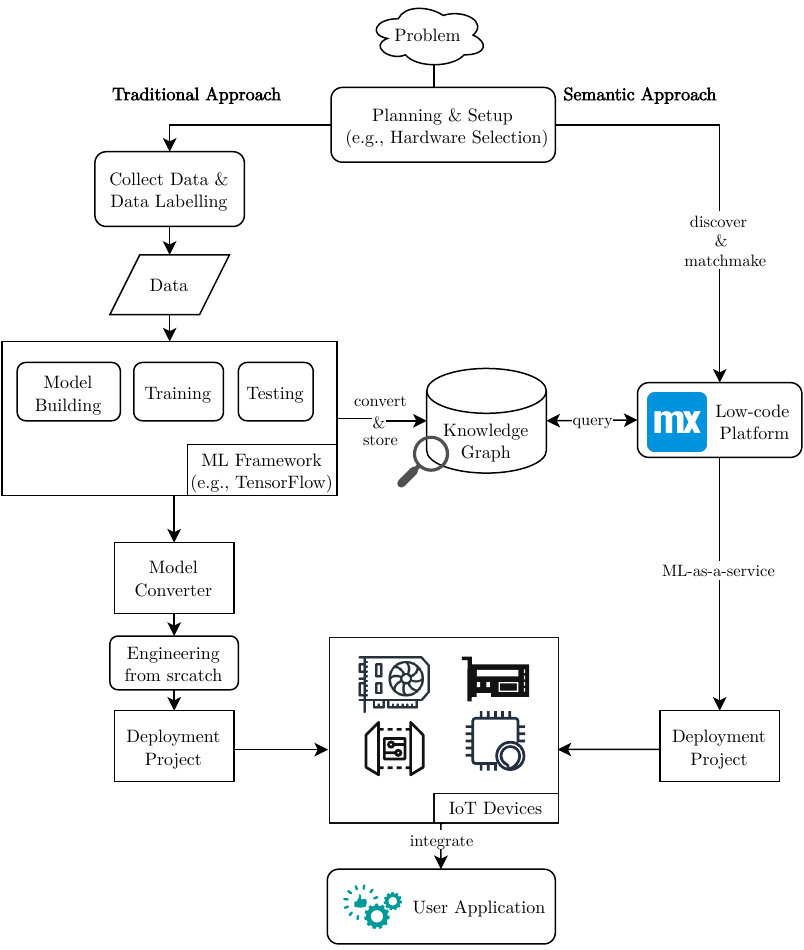}
      \caption{SeLoC-ML: Comparison between the traditional and our semantic approach.}
      \label{seloc_compare_pic}
\end{figure}

Traditionally, a TinyML project begins with planning followed by ML model engineering. ML engineers play a crucial role in model development, from data collection and labeling to model design and training. Afterward, embedded developers optimize and integrate the trained model into an embedded project, ready to be uploaded to the target platform. Finally, software engineers create backend and frontend applications to interact with IoT devices and deliver results to end-users. It is conspicuous that creating TinyML applications from scratch is daunting due to the need for cross-domain expertise and significant engineering effort, not to mention that industries may face the challenge of needing to develop a large number of TinyML applications.

SeLoC-ML offers an integrated solution to streamline the processes. Figure~\ref{seloc_compare_pic} compares the workflows between SeLoC-ML and the traditional approach. SeLoC-ML relies on its semantic system, whose core asset is a KG. One can discover various information in a KG and refine the results by constructing SPARQL queries. Interested readers can find sample queries to demonstrate three case studies in our work~\cite{Ren2022}:

\begin{enumerate}
 \item How do we determine which IoT devices may execute a specific NN? 
 \item Given a device, how do we examine which trained NN model is compatible with it?
 \item How can we benchmark the results based on different metrics and find the most suitable model/device?
\end{enumerate}

Further queries and industrial implementation of SeLoC-ML are also available in our publication and repository. By employing various queries, we can explore all the reusable NN models and IoT devices in the KG and efficiently identify compatible combinations, bypassing the extensive manual work required in the traditional approach.

\begin{figure}[tbp]
      \centering
      \begin{subfigure}[b]{0.4875\textwidth}
         \centering
         \includegraphics[width=\textwidth]{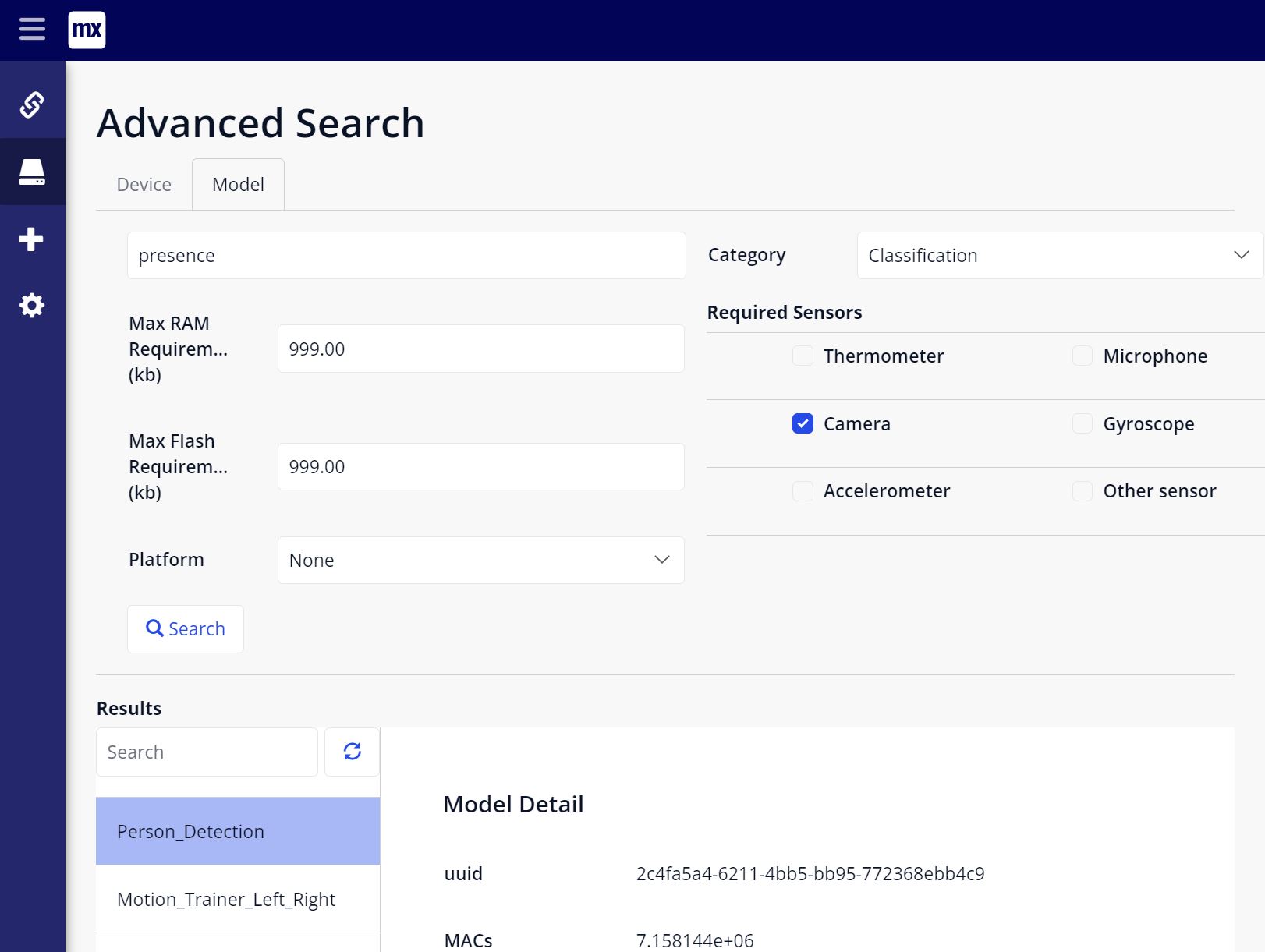}
         \caption{Discovery: Search desired models.}
         \label{matchmaking_results_1}
     \end{subfigure}
     \hfill
     \begin{subfigure}[b]{0.49\textwidth}
         \centering
         \includegraphics[width=\textwidth]{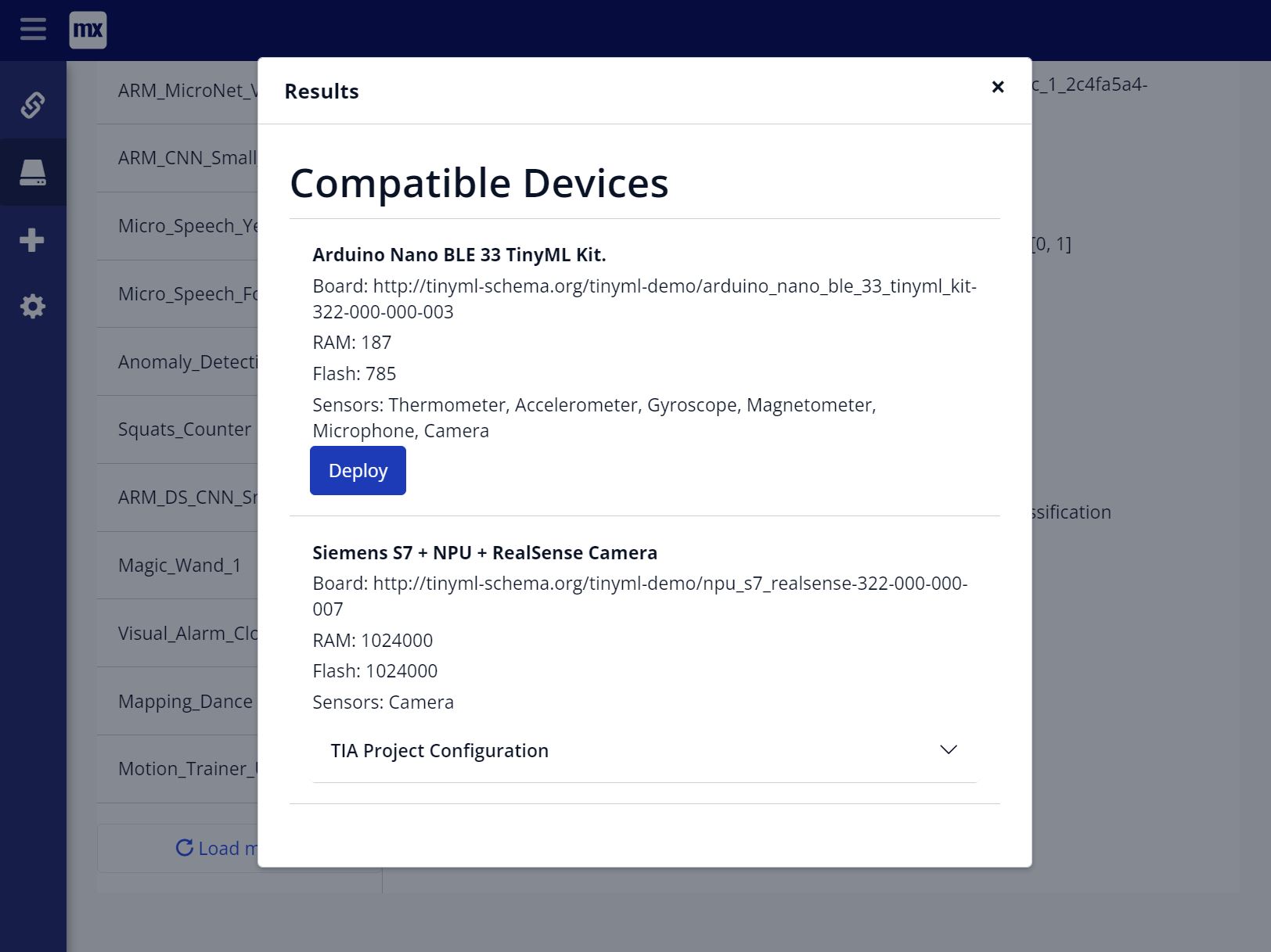}
         \caption{Matchmaking: Find compatible devices.}
         \label{matchmaking_results_2}
     \end{subfigure}
      \caption{SeLoC-ML: Discovery and matchmaking in Mendix.}
\end{figure}



We integrate SeLoC-ML into the Mendix low-code platform, allowing developers to easily navigate and matchmake components as efficiently as browsing the web without the need to construct SPARQL queries. Figure~\ref{matchmaking_results_1} depicts the interface of our Mendix application, where we search for available models using the search term "presence" while applying several filters, such as Required Sensors "Camera." Upon selecting a model from the returned list, Figure~\ref{matchmaking_results_2} illustrates the system's functionality to automatically reason the compatible devices capable of executing the selected model. After matchmaking, various deployment options tailored to the chosen hardware platform become available. 
Users can then quickly prototype their ML applications and generate engineering code with minimum effort, ready for deployment on the hardware.

Now, we quantitatively evaluate SeLoC-ML and the traditional method in terms of engineering effort and error rate for generating TinyML projects, focusing on integrating the trained NN into the project for deployment on the selected hardware.  
The file structure of the TinyML deployment projects includes a shell file for dependency installation on Raspberry Pi and a main Python file for inference execution.
The comparison of engineering effort involves counting Lines of Code (LoC) required for manually programming the project, such as some necessary configuration code.

\begin{table}[tbp]
\centering
\caption{Comparing the engineering effort (LoC) required for each component of deployment projects between the semantic and traditional approach of Application \rom{1} and \rom{2}.}
\label{tab:loc-1}
\begin{tabular}{@{}llll@{}}
\toprule
 & \textit{requirements.sh} & \textit{main.py} & Total  \\ \midrule
\begin{tabular}[c]{@{}l@{}}Handwritten Character Image Classification \\ (traditional approach / semantic approach)\end{tabular} & 7 / 0 & 228 / 20 & 235 / \textbf{20} \\
\begin{tabular}[c]{@{}l@{}}Keyword Spotting Audio Classification \\ (traditional approach / semantic approach)\end{tabular} & 6 / 0 & 337 / 16 & 343 / \textbf{16} \\ \bottomrule
\end{tabular}
\end{table}

Table~\ref{tab:loc-1} presents the results, where SeLoC-ML demonstrates a significant reduction of 91.5\% in LoC compared to the traditional method. SeLoC-ML automatically generates most code by populating predefined templates and parsing key components such as the model binary and program sketch. Additionally, automated code generation reduces programming errors, as validated code should be error-free once generated. SeLoC-ML encourages reusing existing ML models rather than reinventing the wheel since reusability means better scalability and less cost. It is important to note that a few user inputs are still required for tasks like sensor initialization, mapping sensor readings to model input, and customizing system responses based on model output.

\subsection*{Application \rom{2}:  Keyword Spotting Audio Classification}
\addcontentsline{toc}{subsection}{Application \rom{2}:  Keyword Spotting Audio Classification}

The setup of the audio classification application is similar to the image classification application. We showcase the application using the 
Speech Command dataset~\cite{Warden2018}, encompassing 35 distinct keywords (e.g., "left" and "right") with over 1000 audio samples per word. 

\subsubsection*{TinyOL}

Suppose we trained an NN to classify $M=4$ keywords from the dataset. However, the model did not yield the expected performance upon deployment, as shown in Figure~\ref{fig_tinyol_keywords}. We identified the problem: the real-world testing data contains background noise, which is not presented in the training data. In this context, we simulate the original signal samples with background noises by introducing Additive White Gaussian Noise with a signal-to-noise ratio of 10 into the testing data. We incorporate the TinyOL system into the trained model to tackle the issue, enabling on-device post-training on the noisy testing data. After sufficient training iterations, we observed a 4.4\% improvement in model accuracy. Table~\ref{tab:tinyoL_benchmark} provides benchmark results compared with the feature extractor + KNN approach on a Raspberry Pi 4, presenting outcomes similar to those in the last application. For example, TinyOL can achieve greater accuracy improvements while requiring less than one-tenth of the memory compared to batch training. This emphasizes the system's compact footprint and efficient performance. 

\subsubsection*{TinyReptile \& TinyMetaFed}

\begin{table}[tbp]
\centering
\caption{Keyword Spotting
Audio Classification: Benchmark of the meta-learning methods on a Raspberry Pi 4. The results are measured on one device for one round.}
\label{tab:keyword}
\begin{tabular}{@{}llllllll@{}}
\toprule
 &
  Receiving &
  \begin{tabular}[c]{@{}l@{}}Local\\ Training\end{tabular} &
  Sending &
  Total &
  \begin{tabular}[c]{@{}l@{}}Communication\\ Cost\end{tabular} &
  \begin{tabular}[c]{@{}l@{}}Memory \\ Requirement\end{tabular} &
  \begin{tabular}[c]{@{}l@{}}Energy\\ Consumption\end{tabular} \\ \midrule
Reptile     & 3.6 s & 13.4 s & 2.6 s & 19.6 s & 1 * N & 8333 KB & 39 J \\
TinyReptile & 3.6 s & 6.5 s  & 2.6 s & \textbf{12.7 s} & \textbf{1}     & \textbf{1409 KB}  & \textbf{26 J} \\
TinyMetaFed & 1.4 s & 6.3 s  & \textbf{0.8 s} & \textbf{8.5 s}  & \textbf{0.36}  & \textbf{1409 KB}  & \textbf{15 J} \\ \bottomrule
\end{tabular}%
\end{table}

Moving on, we demonstrate how federated meta-learning facilitates rapid customization of keyword commands for end-users. We consider a group of clients, each tasked with classifying four randomly selected keywords. We aim to find a robust NN initialization capable of quickly adapting to new clients' unique conditions. Figure~\ref{fig_meta_keyworkds} presents the training convergence of the three meta-learning methods to the total parameter communicated. The hardware benchmarking results are displayed in Table~\ref{tab:keyword}. Similar to the image classification application, our methods, TinyReptile and TinyMetaFed, improve upon Reptile across most metrics. For instance, TinyMetaFed achieves a 61\% energy saving and 83\% memory reduction compared to Reptile, assuming a batch size of eight in Reptile. We also note substantial time savings with TinyReptile and TinyMetaFed compared to Reptile, with 33\% and 56\%, respectively.

\subsubsection*{SeLoC-ML}

Next, we assess the engineering effort required by SeLoC-ML to generate the TinyML project for this application, comparing it to the traditional approach illustrated in Figure~\ref{seloc_compare_pic}. The results are presented in Table~\ref{tab:loc-1}, focusing on integrating the trained NN into the project for deployment. As this application is also evaluated on Raspberry Pi, it shares the same file structure as the project in the previous application. We observe a higher engineering effort saving of 95.3\% through SeLoC-ML than the last application. This is attributed to the fact that this engineering project requires several audio processing functions, which SeLoC-ML can automatically generate. Also, the generated code should be error-free once the process is validated. Our framework consolidates services into a single tool capable of automating the engineering processes while complying with semantic standards, thus reducing engineering errors and enhancing reliability.

\subsection*{Application \rom{3}:  Smart Building Presence Detection}
\addcontentsline{toc}{subsection}{Application \rom{3}:  Smart Building Presence Detection}

The third application uses environmental sensor readings, time-related information, and room type details to predict room occupancy in a building. We obtained the time series data from 61 rooms in a six-floor commercial building building of Siemens. The sensors have been continuously collecting data in an InfluxDB database since 2021. 

Each room is equipped with four types of IoT sensors: $CO_2$ concentration, room temperature, room air humidity, and infrared motion sensors. Time-related information contains the day of the week, the hour of the day, and holiday information. We categorize the rooms in the building into meeting rooms, office rooms, coffee rooms, and others. We use the motion sensor reading as ground truth for the ML tasks. The motion sensors and $CO_2$ sensors record data every minute, while other sensors are sampled every five minutes.

\subsubsection*{TinyOL}

Suppose we have collected data from a room over a month and used the data to train an NN model for presence detection in that room, a binary ML classification task whose output is a value between 0 (no presence detected) and 1 (presence detected). Later, we deploy the model on testing data, which comes from different rooms or the same room but at different months. Figure~\ref{fig_tinyol_smart_building} displays the training progress, showing that the model's accuracy drops by about 5\% on the testing data. TinyOL mitigates this issue by fine-tuning the model after deployment, as depicted in the figure, resulting in a 2\% model performance improvement. The improvement is not very significant, possibly due to the insufficient numerical precision on the Arduino board for high-resolution gradient updates. As indicated in Table~\ref{tab:tinyoL_benchmark}, the time and energy required for training TinyOL on the Arduino board are similar to those needed for inference, specifically, 0.023~s versus 0.018~s and 0.0056~J compared to 0.0044~J for one iteration. Compared to the KNN method, TinyOL has the advantage of requiring fewer resources for inference and has lower and consistent memory overhead. In IoT scenarios where sensor data is generated continuously, the KNN-based method can become impractical because embedded systems have limited memory capacity for storing extensive training data. However, we observe that the KNN method offers better accuracy in this application, with a result of 89.8\%, compared to 88.4\% achieved by TinyOL. This difference can again be attributed to the Arduino board's limited numerical precision, which impacts TinyOL's training performance. In contrast, KNN is less dependent on high-precision numerical calculations. This finding supports TinyOL's efficiency and highlights embedded devices' limitations in complex computational scenarios.

\subsubsection*{TinyReptile \& TinyMetaFed}

Considering the variability in room construction, sensor placements, and user behaviors across rooms, a global model may perform arbitrarily poorly when applied to different rooms. Federated Meta-learning becomes handy in this scenario, where we treat each room as a client with its own ML classification task for presence detection. This concept allows us to harness the collective knowledge from all the clients to create a robust NN initialization that can quickly adapt to a new deployment environment, be it a different room or the same room but at a different time, with minimum labeled data. Thus, we group data for each ML task based on room numbers. Each task's output is a percentage value ranging from 0 to 1, indicating the occupancy likelihood for the corresponding room.

\begin{table}[tbp]
\centering
\caption{Smart Building Presence Detection: Benchmark of the meta-learning methods on an Arduino Nano BLE 33. The results are measured on one device for one round.}
\label{tab:smart_building_meta}
\begin{tabular}{@{}llllllll@{}}
\toprule
 &
  Receiving &
  \begin{tabular}[c]{@{}l@{}}Local\\ Training\end{tabular} &
  Sending &
  Total &
  \begin{tabular}[c]{@{}l@{}}Communication\\ Cost\end{tabular} &
  \begin{tabular}[c]{@{}l@{}}Memory \\ Requirement\end{tabular} &
  \begin{tabular}[c]{@{}l@{}}Energy\\ Consumption\end{tabular} \\ \midrule
Reptile     & 6.2 s & 8.3 s & 4.2 s & 18.7 s & 1 * N & 22 KB & 5.1 J \\
TinyReptile & 6.2 s & 2.2 s & 4.2 s & \textbf{12.6 s} & \textbf{1}     & \textbf{18 KB} & \textbf{3.9 J} \\
TinyMetaFed & 3.3 s & 2.2 s & 1.4 s & \textbf{6.9 s}  & \textbf{0.39}  & \textbf{18 KB} & \textbf{2.1 J} \\ \bottomrule
\end{tabular}%
\end{table}

Figure~\ref{fig_meta_smart_building} illustrates the training progress of our methods, TinyReptile and TinyMetaFed, compared to Reptile. We provide hardware benchmark results on an Arduino Nano BLE 33 board in Table~\ref{tab:smart_building_meta}. Our methods, TinyReptile and TineMetaFed, achieve comparable accuracy to Reptile while outperforming it in terms of training speed, resource requirements, and communication efficiency. These improvements are similar to those achieved on the Raspberry Pi in the last two applications, for instance, demonstrating at least 73\% and 23\% savings in training time and energy consumption compared to the baseline.

\subsubsection*{SeLoC-ML}

\begin{table}[tbp]
\centering
\caption{Comparing the engineering effort (LoC) required for each component of deployment projects between the semantic and traditional approach of Application \rom{3}.}
\label{tab:loc-2}
\begin{tabular}{@{}llllll@{}}
\toprule
 & \textit{model.h} & \textit{main.cpp} & \textit{input\_handler.h} & \textit{output\_handler.h} & Total \\ \midrule
\begin{tabular}[c]{@{}l@{}}Smart Building Presence Detection\\ (traditional approach / semantic approach)\end{tabular} & 6 / 0 & 126 / 7 & 54 / 48 & 46 / 20 & 232 / \textbf{75} \\ \bottomrule
\end{tabular}
\end{table}

The SeLoC-ML framework exhibits remarkable versatility, accommodating diverse scenarios through its semantic implementation.
Firstly, we outline the C++ embedded project to be generated for this application, which has four engineering components: an input handler responsible for pre-processing sensor data, an output handler for refining model outputs, a binary representation of the ML model, and a main file encompassing variable definitions and program logic.
In Table~\ref{tab:loc-2}, we present the measured engineering efforts for these components, comparing SeLoC-ML against the conventional approach. Notably, SeLoC-ML yields a substantial 67.7\% reduction in engineering effort, primarily attributed to the main cpp file.
In short, SeLoC-ML establishes a unified language for interpreting heterogeneous information in the domain of TinyML, fostering coherent comprehension between human stakeholders and computational systems. Consequently, our framework operates vendor-agnostically and platform-independently, enhancing ecosystem interoperability and transparency.

\section{Conclusions and Future Work}
\label{sec:conclusions}

\begin{figure*}[tbp]
      \centering
      \includegraphics[width=0.99\textwidth]{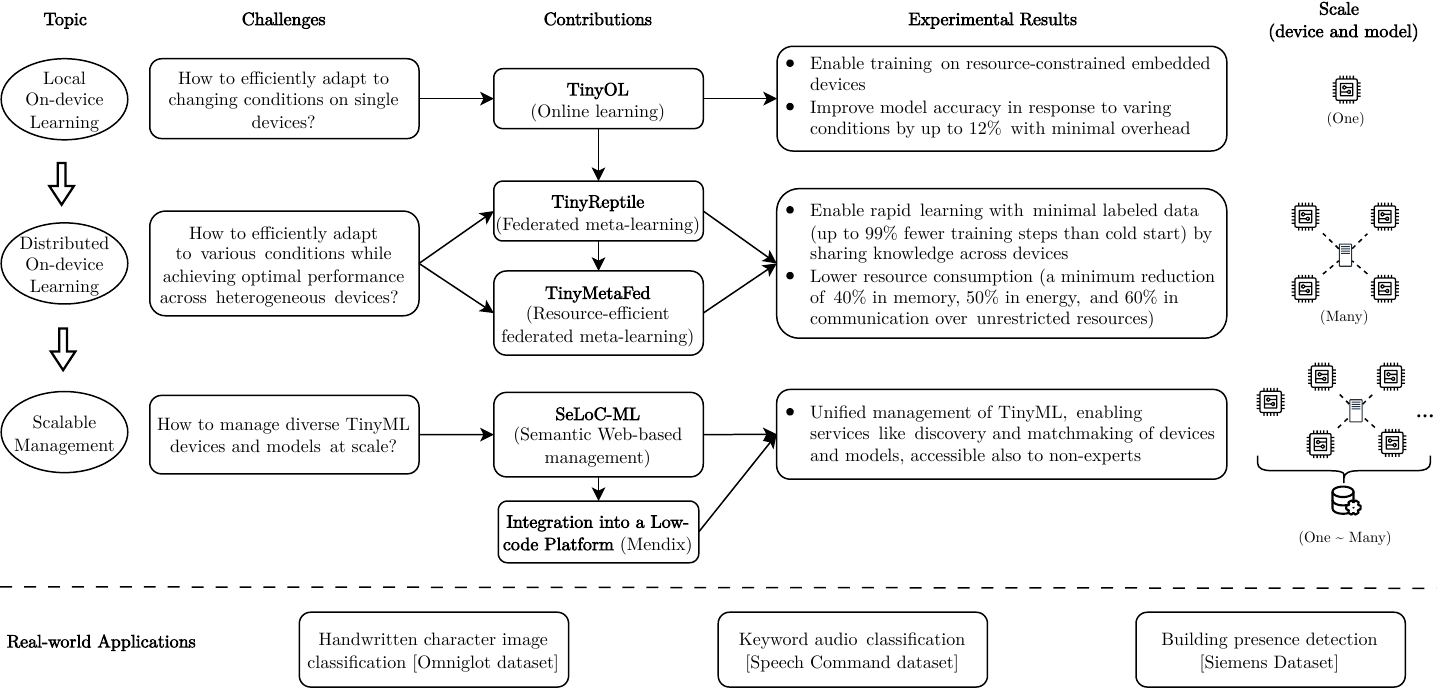}
      \caption{Summary of the challenges, contributions, applications for evaluation, and results of our study, emphasizing their interconnections.}
      \label{summary_diagram}
\end{figure*}

Despite creating TinyML models being quick and straightforward, developing and maintaining TinyML systems in industries presents substantial challenges. Our work identifies three critical challenges in the existing workflow and presents corresponding approaches from our prior research to address them, from small-scale deployment with single devices to mass deployment encompassing distributed devices. Figure~\ref{summary_diagram} summarizes these challenges, our contributions, and experimental results. This work emphasizes their roles from an industrial standpoint throughout the workflow, validates their effectiveness through extensive experiments on novel real-world applications, and offers guidance on leveraging their combined value.

By leveraging the conventional TinyML pipeline, as depicted in Figure~\ref{fig_1}, we construct static models intended only for inference on IoT devices. Nevertheless, real-world conditions constantly change, leading to model performance degradation during deployment. To address this, we propose \textbf{TinyOL} by leveraging online learning, making efficient training on resource-constrained IoT nodes possible. This allows individual models to adapt to the current conditions at runtime for robust performance. However, generalization issues arise in heterogeneous deployment conditions when applying TinyML models to multiple devices while preserving their optimal performance. Fine-tuning the model on each device with TinyOL can exhibit unsatisfactory results and be expensive due to limited labeled data and the considerable number of devices. We introduce \textbf{TinyReptile} for federated meta-learning incorporating online learning, facilitating fast model adaptation with few data. This concept accommodates heterogeneous environmental variations and the demand for customized predictions among distributed devices by enabling knowledge sharing through a central server. Recognizing the need to further reduce resource consumption and enhance privacy, we propose \textbf{TinyMetaFed} as an improvement to TinyReptile. Last but not least, we highlight that the TinyML ecosystem is fragmented, lacking efficient means to manage diverse resources throughout the development. We offer \textbf{SeLoC-ML} to tackle the problem, a user-friendly Semantic Web-based framework \textbf{integrated with the Mendix low-code platform}. SeLoC-ML enables the unified management of TinyML models and devices, providing a range of features, such as knowledge discovery, similarity search, and component matchmaking and reuse, accessible also to individuals without domain expertise. Interested readers can refer to the corresponding publications for a more detailed technical introduction with additional evaluations.

Our empirical results show that TinyOL can improve model accuracy in the field by up to 12\% across three applications. Notably, the energy consumption and training time associated with TinyOL are as low as those incurred during inference, exhibiting minimal computational overhead. Compared with the KNN-based method, our results indicate that TinyOL performs better in terms of most metrics, such as inference overhead and memory usage, even though its accuracy on the Arduino board is slightly lower. This discrepancy suggests that embedded devices may struggle with high-precision computational demands. Exploring ways to achieve similar performance with lower-bit computations could be a promising direction for future research. We observe that the final performance of the fine-tuned models still lags a bit behind the performance achieved in the training dataset, raising the problem of fragile co-adaptation~\cite{Gumbira2018} where the last layers alone may not effectively adapt to the new condition. Furthermore, providing labels on embedded devices can be cumbersome, making TinyOL more practical for unsupervised problems like anomaly detection. Future research could focus on robust co-adaptation and efficient expert feedback. 

TinyReptile and TinyMetaFed demonstrate rapid local adaption, outperforming Reptile by conserving a minimum of 50\% energy, 40\% memory resources, and 60\% communication overhead. Given that only a few local training steps are necessary to achieve good prediction performance, compared to the hundreds or thousands of steps required by cold start (training from scratch), federated meta-learning significantly accelerates local learning, resulting in up to 99\% savings in training steps. Indeed, it is essential to conduct further comparisons of our algorithms with other state-of-the-art approaches and establish best practices for their effective deployment in real-world scenarios. Future research could also explore optimizing hyperparameters to enhance performance and incorporating zero-shot learning to further reduce the requirement for labeled data.

SeLoC-ML can efficiently handle diverse TinyML resources, simplifying the management of TinyML applications. By leveraging the Semantic Web and low-code platforms, we minimize engineering efforts by up to 95\% and decrease error rates by abstracting tedious engineering processes. This promotes the scalability, shareability, and interpretability of TinyML resources. Our goal is to encourage collaboration between the Semantic Web and TinyML communities. Certainly, our approach still needs to be integrated into production progress. And our current implementation does not yet encompass on-device training-related information. This is an area that should be further developed in the next phase. Also, further analysis of other scenarios and platforms, along with collecting feedback, is necessary to advance the robustness and scalability of our system. As part of our future work, provisioning the toolchain and embracing contributions are vital priorities.
 
We aim to streamline the design, deployment, and maintenance of TinyML applications on a scale. While ML code constitutes only a tiny portion of this task, many other essential components in the system need to be addressed for future work. These include system configuration, data acquisition, verification, testing and debugging, process management, serving infrastructure, and use case validation. Fortunately, the research community and industries are paying increasing attention to scaling and operationalizing TinyML. We believe that the future of ML is bright and tiny\footnote{\url{https://pll.harvard.edu/course/future-ml-tiny-and-bright?delta=0}}.

\begin{acks}
This work is partially supported by the NEPHELE (ID: 101070487) and SMARTEDGE (ID: 101092908) projects that have received funding from the European Union’s Horizon Europe research and innovation program.
\end{acks}

\bibliographystyle{ACM-Reference-Format}
\bibliography{sample-base}

\end{document}